\documentclass[journal,twoside,web]{ieeecolor2}
\usepackage{generic}
\usepackage{cite}
\usepackage{amsmath,amssymb,amsfonts}
\usepackage{algorithmic}
\usepackage{graphicx}
\usepackage{booktabs} 
\usepackage{multirow}
\usepackage{array}
\usepackage{rotating}
\usepackage{ragged2e}
\usepackage{hyperref}
\hypersetup{hidelinks,
	colorlinks=true,
	allcolors=black,
	pdfstartview=Fit,
	breaklinks=true}
\usepackage{balance}
\usepackage{colortbl}
\usepackage{algorithm}
\usepackage{algorithmic}

\usepackage{textcomp}
\def\BibTeX{{\rm B\kern-.05em{\sc i\kern-.025em b}\kern-.08em
    T\kern-.1667em\lower.7ex\hbox{E}\kern-.125emX}}
\begin{document}
\title{Extendable Generalization Self-Supervised Diffusion for Low-Dose CT Reconstruction}
\author{Guoquan Wei, Liu Shi, Zekun Zhou, Mohan Li, Cunfeng Wei, Wenzhe Shan, Qiegen Liu \IEEEmembership{Senior Member, IEEE}
\thanks{This work was supported by the National Natural Science Foundation
	of China (Grant: 621220033 and 62201193), Nanchang University Youth Talent
	Training Innovation Fund Project (Grant: XX202506030012), Early-Stage Young Scientific and Technological Talent Training Foundation of Jiangxi Province (Grant: KJ202509220191). (G. Wei is the first author.) (Co-corresponding authors: L. Shi and Q. Liu.)}
\thanks{G. Wei, L. Shi, W. Shan, and Q. Liu are with School of
	Information Engineering, Nanchang University, Nanchang 330031,
	China (email: \{shiliu, shan, liuqiegen\}@ncu.edu.cn, guoquanwei@email.ncu.edu.cn).}
\thanks{Z. Zhou is with School of Mathematics and Computer Sciences, Nanchang University, Nanchang 330031, China (email: zekunzhou@email.ncu.edu.cn).}
\thanks{M. Li and C. Wei are with Jinan Laboratory of Applied Nuclear Science, Jinan 251401, China and also with Institute of High Energy Physics, Chinese Academy of Sciences, Beijing 100049, China (email: \{mohanli, weicf\}@ihep.ac.cn).}
}

\maketitle

\begin{abstract}
\emph{Objective:} To enable dose-extensive generalization using only single-dose projection data for training, this work proposes a novel method of Extendable GENeraLization self-supervised Diffusion (EGenDiff) for low-dose CT reconstruction. \emph{Method:} Specifically, a contextual subdata self-enhancing similarity strategy is designed to provide an initial prior for the subsequent progress. During training, the initial prior is used to combine knowledge distillation with a deep combination of latent diffusion models for optimizing image details. On the stage of inference, the pixel-wise self-correcting fusion technique is proposed for data fidelity enhancement, resulting in extensive generalization of higher and lower doses or even unseen doses. \emph{Results:} EGenDiff requires only LDCT projection data for training and testing. Comprehensive evaluation on benchmark datasets, clinical data, photon counting CT data, and across all three anatomical planes (transverse, coronal, and sagittal) demonstrates that EGenDiff enables extendable generalization multi-dose, yielding reconstructions that consistently outperform leading existing methods. \emph{Significance:} This work mitigates the reliance on paired data, enabling clinically viable generalizability through single-dose training.
\end{abstract}


\section{Introduction}
\label{introduction}

\IEEEPARstart{C}{omputed} tomography (CT) is a clinically established imaging technique whose inherent radiation exposure may increase cancer risk \cite{brenner2007computed,smith2009radiation}. Low-dose CT (LDCT) reduces radiation dose by decreasing tube current, yet introduces severe noise and artifacts that degrade diagnostic quality \cite{sodickson2009recurrent}.

Over the past few decades, LDCT reconstruction has produced numerous commendable results. These studies fall into three main categories: 1) sinusoidal domain data filtering \cite{balda2012ray}, 2) iterative reconstruction \cite{sidky2008image}, and 3) image post-processing techniques \cite{chen2014artifact}. Sinusoidal domain filtering, such as bilateral filtering method \cite{manduca2009projection}, penalized weighted least squares (PWLS) \cite{wang2006penalized}, etc., operates directly on projection data but risk losing frequency information and reducing spatial resolution. Iterative reconstruction is often limited in clinical application due to its high computational cost and complexity. Image post-processing techniques are applied to the reconstructed LDCT images to enhance the image quality, such as block matching based 3D filtering (BM3D) \cite{dabov2007image} demonstrates excellent performance in LDCT imaging.

In recent years, deep learning (DL) has surpassed classical methods in medical imaging applications like traditional CT and photon counting CT (PCCT) \cite{willemink2018photon} reconstruction, where data-driven models are trained to denoise images \cite{shan2019competitive,wang2020deep}. However, current deep learning based methods either use supervised methods that rely on paired data \cite{shen2022mlf,chen2017low,li2025ddoct} or self-supervised methods with paired noise \cite{lehtinen2018noise2noise}, self-supervised learning with batch data similarity \cite{niu2022noise} or unsupervised learning with clean data using diffusion models \cite{garber2024image}. The dominant supervised approach is clinically hampered. Acquiring the necessary real, paired low- and normal-dose data from a single patient is infeasible. This makes self-supervised learning, which trains solely on noisy images, a more practical clinical solution. However, current self-supervised methods still face the challenge of poor generalization when extended to different doses and fields.

Diffusion modeling (DM) \cite{ho2020denoising,dong2025} has recently emerged as a novel and influential explicit generative model that has been widely used in medical image reconstruction with impressive results. Although DM has been shown to achieve high-quality reconstruction \cite{song2020score}, the standard diffusion model still suffers from 1) long sampling time of 1000 steps, difficult to train with high resolution, and difficulty in lightweight deployment \cite{hoogeboom2023simple}. 2) In the absence of real data distribution involved, the results are usually not good. Therefore, a series of methods have emerged. For example, Aali $\emph{et al.}$ \cite{aali2024ambient} first proposed a framework for training on corrupted observation data to solve different forward process inverse problems, and the results are better than training on clean data. Other methods include accelerated samplers \cite{zheng2023dpm}, domain iterative diffusion \cite{liao2024domain}, latent diffusion \cite{nam2024contrastive}, and guided diffusion \cite{hu2023self} have emerged.

In order to better enhance the scalability of generalization, the aim of this work is to explore a low-dose reconstruction method that relies only on a single-dose projection data and does not rely on the distribution of clean data for modeling. To this end, this study proposes extendable generalization diffusion for low-dose CT reconstruction. A contextual LDCT projection domain self-enhancing similarity strategy is designed to adequately remove the noise under the data domain. The results are further optimized in the image domain considering the complexity of noise and incomplete detail recovery. For this purpose, the initial prior is used to deeply combine the knowledge distillation technique with the latent diffusion model to perceive the data consistency and refinement. In the inference stage, the initial prior and LDCT images are introduced as a guide to fully utilize the model effectiveness as well as the matching training process. On this basis, a pixel-level self-correcting fusion is proposed to balance the differences between the bootstrap information, and the technique is skillfully extended to the generalization of the higher doses, lower doses, doses of unseen data, and PCCT data.

In summary, the work of this paper can be described as:
\begin{itemize}
	\item \textbf{Self-Supervised Single Dose Projection Denoising} A novel self-supervised reconstruction model EGenDiff starting from a single dose LDCT projection data is proposed. A self-enhancing similarity strategy for contextual projection sub-data is designed for sufficiently removing the nonlinear noise under the projection domain to provide an effective initial prior.
	\item \textbf{Latent Diffusion for Detail Perception} Mapping the image domain to the latent space, with the help of Gaussian diffusion mechanism and transformer denoising network to fully perceive the details and distill better reconstruction results utilizing initial priors.
	\item \textbf{Extendable  Generalization} EGenDiff exploits the projection domain prior and the LDCT image as a guide in the inference phase, and the pixel-level self-correcting fusion technique improves the balance between the two, facilitating data fidelity and lesion reproduction. The technique is also used for extendable generalization of higher and lower dose, unseen dose, real clinical data and low-energy spectrum data.
\end{itemize}

\section{Related Work}
This section begins with statistical theory-based denoising methods, progresses to approaches leveraging CT data features, and concludes with recent advances in diffusion models for denoising and reconstruction.

\subsection{Denoising Research under Statistical Theory}
\label{subsec1}

Given a LDCT projection measure $y_{ld}$, the inverse problem is to recover to clean data $y_0$. The corresponding image domain is as follows:
\begin{equation}y_{ld}\approx Ax_{0}+\epsilon_{y}, {x_{ld}=\mathcal{R}(y_{ld})=\underbrace{x_{0}+bias}_{\mathcal{R}(Ax_0)}+\underbrace{\epsilon_{x}}_{\mathcal{R}(\epsilon_y)}},\label{eq1}\end{equation}
where $A$ is the projection matrix, $\epsilon_{y}$ is the projection noise, $\mathcal{R}$ is the back-projection operator, and $\epsilon_{x}$ is the reconstruction noise term.

Noise2Noise \cite{lehtinen2018noise2noise} eliminated the need for clean data or paired training samples by utilizing pairs of independent noisy realizations $x_{ld}$ and $x_{ld}^\prime$, thereby achieving an effect equivalent to supervised learning with clean reference $x_0$ during training. The statement is as follows:
\begin{equation}
	\begin{split}
		\min_\theta \left\{ \mathbb{E} \| f_\theta(x_{ld}^{\prime}) - x_{ld} \|^2 \right\} & \\
		\approx \min_\theta \left\{ \mathbb{E} \| f_\theta(x_{ld}^{\prime}) - x_0 \|^2 \right. & \left. + \, \mathbb{E} \| x_{ld} - x_0 \|^2 \right\},
		\label{eq3}
	\end{split}
\end{equation}
where $f_\theta(\cdot)$ represents a neural network driven by $\theta$. However, this method requires two independent noise measurements of the same information and is not strictly self-supervised.

On this basis, Noise2Self \cite{batson2019noise2self} extended J-invariance theory by focusing on the input $x_{ld}$. Noiser2Noise \cite{moran2020noisier2noise} constructsed training pairs by adding further noise to noisy images, Millard $\emph{et al.}$ \cite{millard2023theoretical} proposed a variant to address the statistical bias caused by variable density sampling in MRI. Yaman $\emph{et al.}$ \cite{yaman2020selfsupervised,yaman2021zeroshot} divided the observation data into mutually exclusive sets for research and conducted a zero-shot self-supervised study applied to MRI, but these were not applicable to the imaging characteristics and clinical scenarios of CT. Neighbor2Neighbor \cite{huang2022neighbor2neighbor} offered an efficient self-supervision approach using adjacent subsamples from a single noisy image. These methods often performed poorly in CT due to the spatial correlation introduced by the reconstruction process.

Consequently, CT-specific noise optimization has been explored. Niu \emph{et al.} \cite{niu2022noise} used patient-specific similar data as denoising targets, though random selection limited generalization. Wang \emph{et al.} \cite{wang2024svb} improved the selection strategy and introduced a visual blind-spot technique. These batch-based methods require aligned Z-axis CT data. Other studies leverage CT imaging principles: Wu \emph{et al.} \cite{wu2023unsharp} applied guided filtering to unsharp CT image structures, while Proj2Proj \cite{unal2024proj2proj} demonstrated self-supervised denoising directly in projection space, validating denoising starting from the projection domain.

\begin{figure*}[!h]
	\centerline{\includegraphics[width=\linewidth]{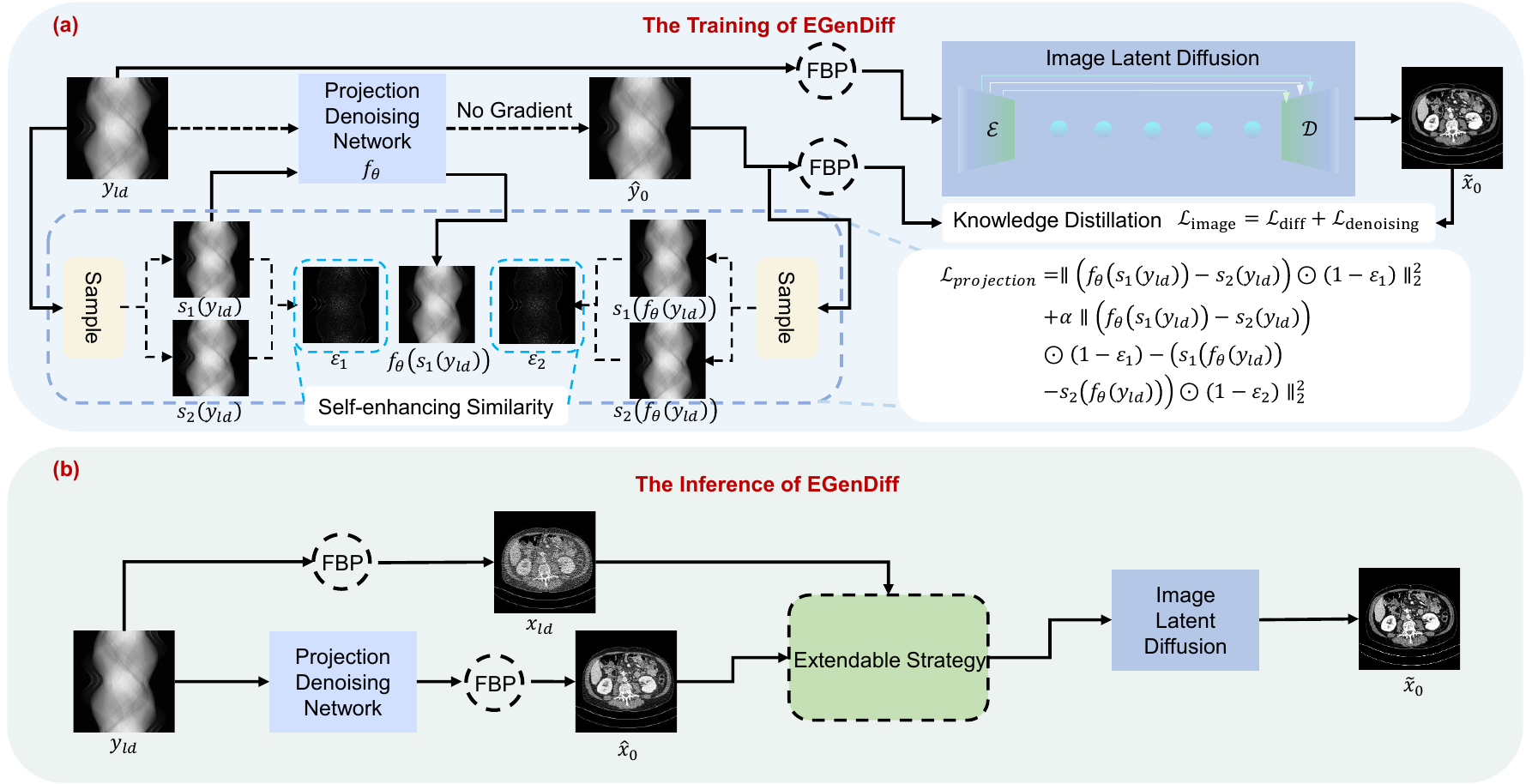}}
	\caption{General Architecture of EGenDiff Training and Inference. (a) The training of EGenDiff. (b) The inference of EGenDiff.}
	\label{fig2}
\end{figure*}
Several similar studies have recently emerged \cite{mansour2023zero-shot,zhao2023sample2sample,liu2022similarity,shi2024zero-ig,aali2025robust,kim2021noise2score,wu2024mu}, but unlike them, this work introduces self-enhancing similarity during CT projection domain training, actively ignoring regions with significant differences, making the model more stable and the theory more complete. This lays a solid foundation for subsequent image domain diffusion refinement and extendable generalization multi-dose processing. Appendix A provides a detailed explanation of this.
\subsection{Diffusion Models}
The diffusion model, as a generative model obeying a Markov chain, is designed to learn an approximate distribution of the data in the forward noise addition and inverse denoising process. The forward process is constructed by adding noise to the given target data:
\begin{equation}q(x_t|x_0)=\mathcal{N}\left(x_t;\sqrt{\bar{\alpha}_t}x_0,(1-\bar{\alpha}_t)\boldsymbol{I}\right),\label{eq4}\end{equation}
\begin{equation}x_t=\sqrt{\bar{\alpha}_t}x_0+\sqrt{1-\bar{\alpha}_t}\epsilon,\epsilon\sim\mathcal{N}(0,\boldsymbol{I}),\label{eq5}\end{equation}
where ${\bar{\alpha}}_t$ is a hyperparameter that controls the noise level, and the forward process adds noise up to $x_T\sim\mathcal{N}(0,\boldsymbol{I})$. The backward process then progressively predicts from $x_T$ until it approaches $x_0$.
\begin{equation}p_\theta(x_{t-1}|x_t){:}=\mathcal{N}(x_{t-1};\boldsymbol{\mu}_\theta(x_t,t),\sigma_t^2\boldsymbol{I}),\label{eq6}\end{equation}
\begin{equation}\boldsymbol{\mu}_\theta(x_t,t)=\frac{1}{\sqrt{\alpha_t}}\left(x_t-\frac{1-\alpha_t}{\sqrt{1-\bar{\alpha}_t}}\epsilon_\theta(x_t,t)\right),\sigma_t^2=1-\alpha_t,\label{eq7}\end{equation}
where $\epsilon_\theta$ is a noise predictor driven by a neural network.

Recent extensions of diffusion models include the score-based generative model by Song \emph{et al.} \cite{song2020score}, which utilized stochastic differential equations (SDEs). Gao \emph{et al.} \cite{gao2023corediff} developed CoreDiff, a cold diffusion-based model preserving physical degradation properties, later extended to the dual-domain model NEED \cite{gao2025noise}, which incorporates offset Poisson diffusion in the projection domain and dual-guided diffusion in the image domain for improved reconstruction. Liao \emph{et al.} \cite{liao2024domain} designed a dual-domain denoising model that enhances sampling efficiency through an iterative partial diffusion process. Yang \emph{et al.} \cite{yang2025lfdt} proposed a latent feature-guided diffusion model applicable to multimodal image fusion tasks. Li \emph{et al.} \cite{li2025prompt} proposed Prompt-SID, which introduces structural cue learning and a scale replay mechanism in latent space to bridge resolution gaps for self-supervised denoising.

\section{Method}
This section introduces a projection domain denoising model that utilizes self-enhancing similarity of contextual subdata, followed by an image domain latent information distillation stage in the latent space. Furthermore, a pixel-wise self-correcting fusion mechanism and an extendable generalization strategy for multi-dose reconstruction are detailed, and their overall framework is shown in Fig. \ref{fig2}. The algorithm flow is as shown in Algorithm \ref{alg:A1}.
\subsection{Projection Domain Contextual Sub-data Denoising}
\begin{figure}[!ht]
	\centerline{\includegraphics[width=\linewidth]{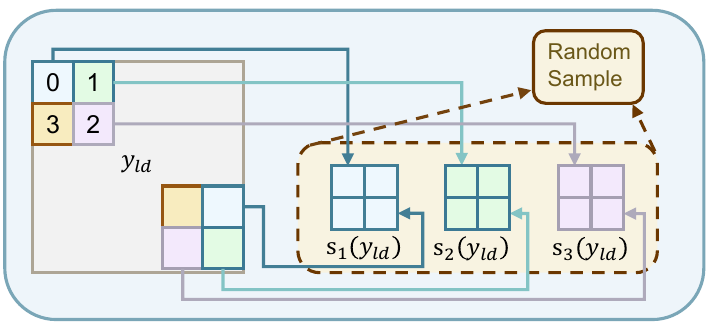}}
	\caption{Description of random sampling of contextual projection data.}
	\label{fig3}
\end{figure}

At present, the noise generated in the LDCT scanning process mainly comes from the quantum noise caused by photon fluctuation and the electronic noise caused by the equipment itself. And the image domain data under the filtered backprojection is prone to bring stronger spatial correlation. According to Eq. \eqref{eq3}, following the noise-to-noise principle can realize that targeting noise is equivalent to recovering to a clean image. To this end, this study proposes a randomized contextual subdata sampling strategy $s(\cdot)$. Firstly, the noise-containing projection $y_{ld}$ is divided into $H_y/2\times W_y/2$ chunks, each chunk contains four data points, and one point is randomly selected among the four data points $ i$, the remaining three are the lower data point $i-1$ or the upper data point $i+1$, $i+2$ of that point, when $i=0$ or $i=3$ there is only the upper data point or the lower data point. As shown in Fig. \ref{fig3}, the selected data points are divided into different sampling data. The random selection strategy can be described as:
\begin{equation}\left( s_{i}(y_{ld}), s_{j}(y_{ld}) \right) \sim \mathcal{U} \left\{ s_{1}(y_{ld}), s_{2}(y_{ld}), s_{3}(y_{ld}) \right\}, \quad i \neq j,\label{eq9}\end{equation}
where $(s_{i}(\cdot),s_{j}(\cdot))\sim\mathcal{U}(\cdot)$ represents the sub-data after random sampling of the context.

According to lemma 1 of \cite{lehtinen2018noise2noise}, and the proof of \cite{zhussip2019extending}. Assuming that noisy image pairs $a$ and $b$ are conditionally independent given a clean image $x_0$, and that the gap between the two is defined as $\varepsilon:=\mathbb{E}_{b|x_0}(b)-\mathbb{E}_{a|x_0}(a)$, the relationship between the noisy data pairs and the supervised training loss is defined as:
\begin{equation}
	\begin{aligned}
		\mathbb{E}_{x_0,a}\parallel f_\theta(a)-x_0\parallel_2^2 &=\mathbb{E}_{x_0,a,b}\parallel f_\theta(a)-b\parallel_2^2\\ &-\sigma_b^2+2\varepsilon\mathbb{E}_{x_0,a}(f_\theta(a)-x_0),\label{eq10}
	\end{aligned}
\end{equation}
where $\sigma_z^2$ is the variance of $b$, when $\varepsilon\rightarrow0$, optimizing $\mathbb{E}_{x_0,a,b}\parallel f_\theta\left(a\right)-b\parallel_2^2$ is equivalent to optimizing the supervised loss. Similarly, $s_1\left(y_{ld}\right)$ and $s_2\left(y_{ld}\right)$ are context pixels in $y_{ld}$, and there is always a certain similarity.

To enhance similarity in the data and provide more stable approximation supervision loss, this study introduces a self-enhancing similarity strategy that dynamically excludes regions with significant disparities between contextual sub-data. The term pairs $(s_1\left(y_{ld}\right), s_2\left(y_{ld}\right))$ and $(s_1\left(f_\theta\left(y_{ld}\right)\right), s_2\left(f_\theta\left(y_{ld}\right)\right))$ are defined to satisfy $\varepsilon_1=\mathbb{E}_{y_{ld}|y_0}\left(s_1\left(y_{ld}\right)\right)-\mathbb{E}_{y_{ld}|y_0}(s_2\left(y_{ld}\right))$ and $\varepsilon_2=\mathbb{E}_{y_{ld}|y_0}\left(s_1\left(f_\theta\left(y_{ld}\right)\right)\right)-\mathbb{E}_{y_{ld}|y_0}(s_2\left(f_\theta\left(y_{ld}\right)\right))$. After extracting differential regions, the model dynamically propagates this result to the supervision loss calculation. The objective function can be defined as:
\begin{equation}
	\begin{aligned}
		\min_{\theta} \;&\mathbb{E}_{y_{0},y_{ld}}\parallel(f_\theta(s_1(y_{ld}))-s_2(y_{ld}))\odot(1-\varepsilon_{1})\parallel_2^2 \\
		& +\alpha\mathbb{E}_{y_{0},y_{ld}}\parallel(f_\theta(s_1(y_{ld}))-s_2(y_{ld}))\odot(1-\varepsilon_{1})\\&-(s_1(f_\theta(y_{ld}))-s_2(f_\theta(y_{ld})))\odot(1-\varepsilon_{2})\parallel_2^2,\label{eq11}
	\end{aligned}
\end{equation}
where $\odot$ represents element-by-element multiplication and $\alpha$ is empirically set to 0.02. For detailed information about Eq. \eqref{eq11}, please refer to the Appendix. A.

The initial denoised projection data ${\hat{y}}_0$ can be obtained after the above processing. The initial reconstructed image domain ${\hat{x}}_0$ can be described as:
\begin{equation}
	\begin{aligned}
		\hat{x}_0=\mathrm{FBP}(\hat{y}_0).\label{eq12}
	\end{aligned}
\end{equation}
\subsection{Latent Diffusion Image Refinement}
While projection-domain denoising significantly reduces noise, the initial reconstruction $\hat{x}_0$ often exhibits residual blurring and incomplete noise removal. Further same processing may introduce error accumulation due to strong spatial correlations in CT images, leading to deviated optimization trajectories and information degradation. To address this, we introduce a latent diffusion model (LDM)-based denoising framework that leverages Gaussian diffusion in latent space to learn uncorrupted features for refined image enhancement.

Conventional LDMs require clean-image pretraining for task-specific encoders, which contradicts self-supervised denoising from a single projection. To resolve this, we adopt the U-Net-based encoder-decoder structure from \cite{yang2025lfdt}, which effectively maps images into a lower-dimensional latent space while preserving high-frequency noise patterns. Based on formulas Eqs. \eqref{eq4}–\eqref{eq6}, the diffusion model is defined as follows:
\begin{equation}
	\begin{aligned}
		x_{ld}^L=\mathcal{E}(x_{ld}),\label{eq13}
	\end{aligned}
\end{equation}
\begin{equation}
	\begin{aligned}
		\tilde{x}_0=D(x_0^L),\label{eq14}
	\end{aligned}
\end{equation}
where $\mathcal{E}$ represents the pixel-level encoder used to map $x_{ld}$ to the latent space variable $x_{ld}^L$. $x_0^L$ represents the latent space variable to be decoded after denoising. $\mathcal{D}$ is then the decoded compression variable $x_0^L$ for the final denoised CT image ${\widetilde{x}}_0$.

To leverage the intermediate representation of the initial reconstruction ${\hat{x}}_0$, this study integrates knowledge distillation into the LDM framework. 1) The pre-trained projection-domain model yields ${\hat{x}}_0$ as a representation-constrained prior. 2) $x_{ld}$ is compressed into a low-dimensional latent space via Eq. \eqref{eq13}, then progressively degraded to Gaussian noise $x_T$ via Eq. \eqref{eq4}. 3) A transformer-based noise prediction network $\epsilon_\theta\left(\cdot\right)$ is trained with Eq. \eqref{eq6} to focus on low-frequency information for inverse denoising, generating $x_0^L$, which is subsequently decoded via Eq. \eqref{eq14}. 4) Iterative training, using ${\hat{x}}_0$ as a constrained prior, distills $\tilde{x}_0$, which has more complete feature information. The reverse process uses only 5 sampling steps, ensuring high efficiency. Therefore, the loss in this stage contains diffusion loss and denoising loss:
\begin{equation}
	\begin{aligned}
		\mathcal{L}_{\mathrm{image}}=\mathcal{L}_{\mathrm{diff}}+\mathcal{L}_{\mathrm{denoising}},\label{eq15}
	\end{aligned}
\end{equation}
where $\mathcal{L}_\mathrm{diff}$ represents minimizing the difference between the denoising network prediction noise $\epsilon_\theta\left(x_t,t\right)$ and the initial sampling noise $\epsilon$.

The denoising loss is designed to exploit knowledge distillation for better fine-grained generation. The definition is as follows:
\begin{equation}
	\begin{aligned}
		&\mathcal{L}_{\mathrm{denoising}}  =\beta\mathcal{L}_{\mathrm{ssim}}+\gamma\mathcal{L}_{\mathrm{grad}}+\eta\mathcal{L}_{\mathrm{l1}} \\
		& =\beta\left(1-\mathrm{ssim}(\mathcal{D}(x_{t}^{L}),x_{0})\right)
		+\gamma\frac{1}{H_{x}W_{x}}\||\nabla\hat{x}_{0}|-|\nabla D(x_{t}^{L})|\|_{1} \\
		& +\eta\frac{1}{H_{x}W_{x}}\|\hat{x}_{0}-\mathcal{D}(x_{t}^{L})\|_{1},
	\end{aligned}\label{eq17}
\end{equation}
where $\beta$, $\gamma$, $\eta$ represent the weights of each loss, which are empirically set to $1.0$, $2.0$, $0.5$. $\nabla$ represents the Sobel-based gradient operator. $x_t^L$ represents a certain latent state that is not fully denoised.
\begin{figure}[!ht]
	\centerline{\includegraphics[width=0.9\linewidth]{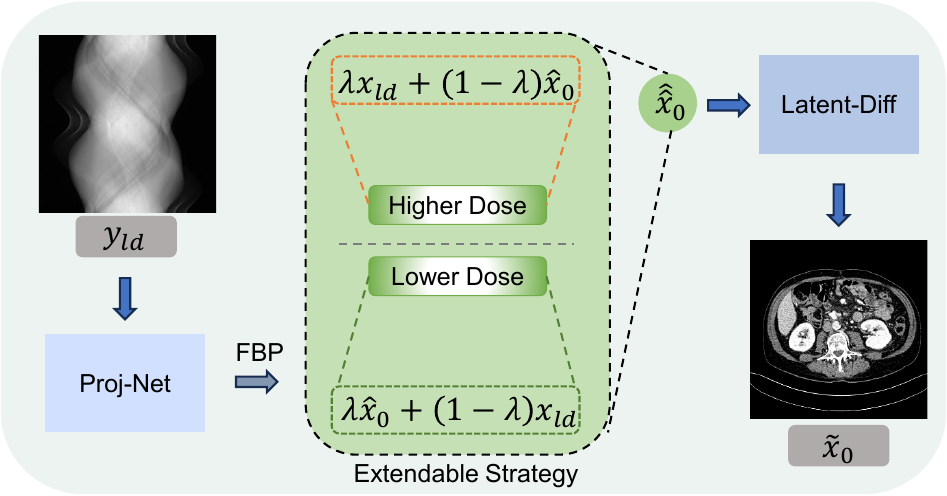}}
	\caption{Extendable higher and lower dose generalization strategy. Different strategies are employed for higher and lower doses. When the noise level is greater than $25\%$ dose, a lower dose strategy is used for generalization; when the noise level is less than $25\%$ dose, a higher dose strategy is used for generalization.}
	\label{fig4}
\end{figure}
\subsection{Extendable Generalization for Higher and Lower dose}
\subsubsection{Pixel-level self-correcting fusion reconstruction}{
A trained projection domain model and an image domain model are used for LDCT dual-domain reconstruction. When the dual-domain model cascade is performed for reconstruction, if the initial reconstruction ${\hat{x}}_0$ in the projection domain is used as the starting point of the image domain stage, it may have the effect of over-suppressing the noise and destroying the medical lesions. To address this, we introduce a weighted fusion of the original LDCT image $x_{ld}$ and the initial reconstruction $\hat{x}_0$ to guide the image-domain reconstruction. This approach compensates for information lost in the projection domain while avoiding the introduction of excessive noise and artifacts. Unlike previous methods that rely on fixed weights or scalar adaptive weights, our method accounts for the non-uniform spatial distribution of noise and anatomical features through pixel-level adaptive weighting.
	
Specifically, the fusion process is guided by two confidence measures: an edge confidence map $C_e$ derived from ${\hat{x}}_0$ and a noise confidence map $C_n$ estimated from $x_{ld}$. The description is as follows:
\begin{equation}
		\begin{array}
			{c}{C_{e}=\sigma(k_{2}(G_{\hat{x_0}}-\tau_{e})),} 
			{C_{n}=\sigma(k_{1}(N_{x_{ld}}-\tau_{n})),}
		\end{array}\label{eq18}
\end{equation}
\begin{equation}
		\begin{aligned} \lambda=\mathcal{F}(C_e, C_n, N_{x_{ld}}^{25\%}), \end{aligned}\label{eq19}
\end{equation}	
where $C_e$ and $C_n$ are the edge confidence and noise confidence. $\sigma(z)=\frac{1}{1+e^{-z}}$ and $\tau_{e}=\mathrm{sort}(G_{\hat{x_0}},0.8), \tau_{n}=\mathrm{sort}(N_{x_{ld}},0.8).$ $G_{{\hat{x}}_0}$ is the gradient magnitude and $N_{x_{ld}}$ is the noise level estimate. $\lambda\in\mathbb{R}^{B_x\times C_x\times H_x\times W_x}$ exists as pixel-level weights. So the image into the image domain is represented as:
\begin{equation}
		\begin{aligned} \hat{\hat{x}}_{0}=\underbrace{\lambda x_{ld}}_{\mathrm{LDCT~term}}+\underbrace{(1-\lambda)\hat{x}_{0}}_{\text{middle term}}, \end{aligned}\label{eq20}
\end{equation}
where $\lambda$ must not be all 1, otherwise this is contrary to the original purpose of cascading dual-domain denoising, ${\hat{\hat{x}}}_0$ is the image to be denoised into the image domain pre-training model, and then denoised by the Eq. \eqref{eq17} pre-trained model to get our final reconstructed image $\tilde{x}_{0}$.}
\begin{figure}[!htbp]
	\centerline{\includegraphics[width=\columnwidth]{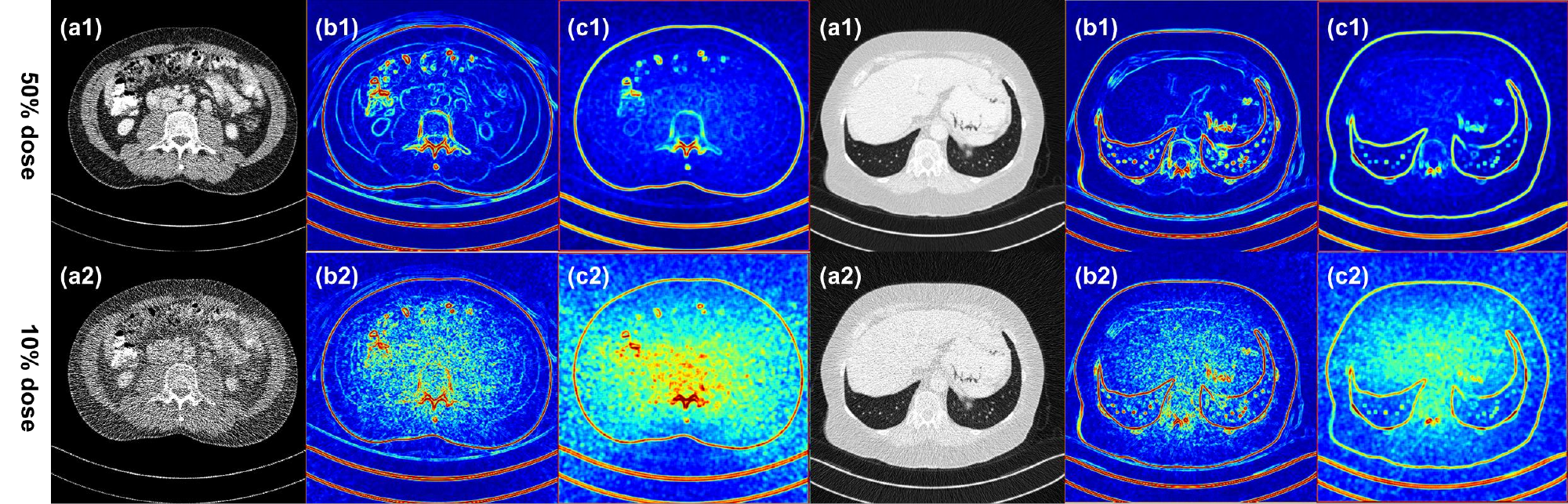}}
	\caption{\{(a1), (a2)\}: noisy images, \{(b1), (b2)\}: the edges extracted from $\hat{x}_{0}$, and \{(c1), (c2)\}: the noise level estimated from $x_{ld}$.}
	\label{fig14}
\end{figure}
\subsubsection{Extendable strategy of higher and lower dose generalization}{Define $\mathcal{F}(\cdot)$ as the key function for extendable generalization, with $N_{x_{ld}}^{25\%}$ representing the noise level at $25\%$ dose. Currently, the dose distribution of LDCT is generally between $5\%$ and $50\%$ dose, with $25\%$ as the boundary, $5\%-25\%$ is the lower dose, and $25\%-50\%$ is the higher dose. The definition is as follows:
\begin{equation}
		\mathcal{F}(C_e,C_n,N_{x_{ld}}^{25\%})=\left\{\begin{matrix} \lambda_{higher} & S\leq S_0  \\ \lambda_{lower} & S>S_0 & \end{matrix},\right.
\end{equation}
where $S=\mathrm{sum}(N_{x_{ld}})$, $S_0=\mathrm{sum}(N_{x_{ld}}^{25\%})$, $\lambda_{higher}=\mathrm{Clip}(C_n+C_e,0,1),\lambda_{lower}=1-\mathrm{Clip}(C_n+C_e,0,1)$, $\mathrm{sum}\left(\cdot\right)$ denotes pixel-wise noise level summation. Different noise levels determine whether $\lambda$ tends to favor $x_{ld}$ or $\hat{x}_{0}$. Detailed description is shown in Fig. \ref{fig4} and Fig. \ref{fig14}. Furthermore, the term "Extendable" should be interpreted not as manual fine-tuning, but as a system-level adaptation. Specifically, the determination of $N_{x_{ld}}^{25\%}$ for a device defines a configuration that permits adaptive, generalized reconstruction for a range of low-dose conditions.
\begin{algorithm}
		\caption{Training and Inference Process}
		\label{alg:A1}
		\begin{algorithmic}[1]
			\REQUIRE Single LDCT projection data $y_{ld}$, designed $s(\cdot)$, total time steps $T$ and sampling steps $T_L$
			\ENSURE Well-trained models $f_\theta$, $\mathcal{E}$, $\mathcal{D}$ and $\epsilon_\theta$; Reconstructed image $\tilde{x}_0$
			\\
			\STATE \textbf{Training Phase:}
			\REPEAT
			\STATE Calculate $s_{1}(y_{\mathrm{ld}})$, $s_{2}(y_{\mathrm{ld}})$ by Eq. \eqref{eq9}
			\STATE Update $f_\theta$ by Eq. \eqref{eq11}
			\STATE Update $\mathcal{E}$, $\mathcal{D}$, $\epsilon_\theta$ by Eq. \eqref{eq15}
			\UNTIL{converged}
			\\
			\STATE \textbf{Inference Phase:}
			\STATE $\hat{y}_0 \leftarrow f_\theta(y_{ld})$
			\STATE Calculate $\hat{x}_0$ by Eq. \eqref{eq12}
			\STATE Calculate $\hat{\hat{x}}_{0}$ by Eq. \eqref{eq20}
			\STATE $\hat{\hat{x}}_{0}^L \leftarrow \mathcal{E}(\hat{\hat{x}}_{0})$
			\FOR{$t=T,T-(T/T_L), \cdots, 1$}
			\STATE Calculate $\hat{\hat{x}}_{t-1}^L$ by Eq. \eqref{eq6}
			\ENDFOR
			\STATE Calculate $\tilde{x}_0$ by Eq. \eqref{eq14}
		\end{algorithmic}
\end{algorithm}
\begin{table*}
	\centering
	\setlength{\extrarowheight}{0pt}
	\addtolength{\extrarowheight}{\aboverulesep}
	\addtolength{\extrarowheight}{\belowrulesep}
	\setlength{\aboverulesep}{0pt}
	\setlength{\belowrulesep}{0pt}
	\caption{Quantitative results (mean±std) on 25\% dose level from the Mayo 2016 dataset and the LIDC-IDRI dataset. The best results are highlighted in bold.}
	\label{table1}
	\resizebox{\linewidth}{!}{%
		\begin{tabular}{ccccccc|cccccc} 
			\toprule
			\multirow{2}{*}{Methods}                       & \multicolumn{2}{c}{}                      & \multicolumn{2}{c}{Mayo Dataset}          & \multicolumn{2}{l|}{}                     & \multicolumn{2}{c}{}                      & \multicolumn{2}{c}{LIDC-IDRI Dataset}     & \multicolumn{2}{l}{}                       \\ 
			\cline{2-13}
			& PSNR $\uparrow$     & SSIM(\%) $\uparrow$ & RMSE $\downarrow$   & FSIM(\%) $\uparrow$ & VIF(\%) $\uparrow$  & NQM $\uparrow$      & PSNR $\uparrow$     & SSIM(\%) $\uparrow$ & RMSE $\downarrow$   & FSIM(\%) $\uparrow$ & VIF(\%) $\uparrow$  & NQM $\uparrow$       \\ 
			\hline
			\textbf{Baseline/Tradition}                    & \multicolumn{12}{l}{}                                                                                                                                                                                                                                                  \\ 
			\hline
			\rowcolor[rgb]{0.753,0.753,0.753} FBP          & 31.89±2.58          & 70.00±10.27         & 53.17±15.93         & 94.74±2.73          & 24.60±7.51          & 28.06±2.89          & 36.39±3.43          & 84.28±9.91          & 32.73±14.90         & 96.84±2.30          & 37.10±13.74         & 31.35±2.69           \\
			TV                                             & 38.21±2.47          & 94.30±2.49          & 24.74±9.21          & 95.31±1.31          & 61.21±13.52         & 22.03±2.68          & 40.05±1.31          & 96.03±0.82          & 20.03±3.61          & 94.68±3.10          & 62.10±9.80          & 28.14±1.51           \\ 
			\hline
			\textbf{Self-supervised}                       & \multicolumn{12}{l}{}                                                                                                                                                                                                                                                  \\ 
			\hline
			\rowcolor[rgb]{0.753,0.753,0.753} B2U          & 32.83±2.39          & 73.41±9.69          & 47.44±13.66         & 96.02±2.04          & 26.29±8.17          & 26.86±2.55          & 36.78±3.61          & 85.22±9.80          & 31.58±13.07         & 97.10±2.17          & 39.29±14.94         & 30.02±2.59           \\
			Neighbor2Neighbor                              & 39.02±2.35          & 92.41±3.92          & 23.25±6.79          & 98.22±0.81          & 53.37±11.39         & 27.47±2.58          & 41.59±2.40          & 96.01±2.61          & 17.32±5.28          & 98.68±0.63          & 64.41±13.60         & 30.65±2.38           \\
			\rowcolor[rgb]{0.753,0.753,0.753} Noiser2Noise & 38.22±2.53          & 92.52±5.24          & 25.78±9.42          & 97.05±0.63          & 61.58±12.84         & 24.28±1.15          & 38.34±0.95          & 96.15±0.97          & 24.33±2.69          & 96.91±0.69          & 67.62±4.61          & 21.39±0.79           \\
			Prompt-SID                                     & 36.71±2.51          & 87.21±6.03          & 30.47±9.20          & 98.05±0.99          & 41.72±10.78         & 28.18±2.58          & 40.43±2.77          & 94.05±3.97          & 20.04±6.78          & 98.86±0.70          & 56.37±14.73         & 31.42±2.32           \\
			\rowcolor[rgb]{0.753,0.753,0.753} Noise2Sim    & 39.32±1.42          & 95.68±0.92          & 21.92±3.84          & 97.86±0.47          & 65.04±5.95          & 26.18±2.87          & 36.70±2.33          & 95.99±0.71          & 30.36±9.17          & 97.21±0.71          & 64.39±7.62          & 20.68±3.63           \\
			AdaReNet                                       & 40.80±3.56          & 96.47±0.32          & 20.44±13.33         & 98.43±0.13          & 72.41±6.04          & 28.21±3.29          & \textbf{43.47±0.64} & 97.90±0.55          & 14.56±2.47          & \textbf{99.14±0.12} & 76.34±2.79          & \textbf{32.33±1.21}  \\ 
			\hline
			\textbf{Unsupervised}                          & \multicolumn{12}{l}{}                                                                                                                                                                                                                                                  \\ 
			\hline
			\rowcolor[rgb]{0.753,0.753,0.753} IPDM         & 41.32±0.99          & 96.78±0.80          & 17.81±3.61          & 98.45±0.43          & 72.90±4.80          & 28.07±2.28          & 42.30±0.74          & 97.84±0.55          & 15.39±1.36          & 98.36±0.36          & 75.30±4.76          & 30.32±1.30           \\ 
			\hline
			\textbf{\textbf{Self-supervised}}              & \multicolumn{12}{l}{}                                                                                                                                                                                                                                                  \\ 
			\hline
			EGenDiff (\textbf{ours})                       & \textbf{42.41±1.21} & \textbf{96.90±0.83} & \textbf{15.30±2.21} & \textbf{98.71±0.33} & \textbf{73.64±5.44} & \textbf{28.25±2.35} & 42.85±0.77          & \textbf{97.95±0.40} & \textbf{14.46±1.38} & 98.95±0.23          & \textbf{76.45±3.79} & 31.54±2.11           \\
			\bottomrule
		\end{tabular}
	}
\end{table*}
\section{Experiments}
This section focuses on the dataset used in this study, the simulation approach, the experimental details, and the evaluation metrics. Subsequently, the proposed method is thoroughly evaluated by quantitatively and qualitatively comparing it with the current state-of-the-art denoising methods. Next, the extendable generalization of EGenDiff will be validated using real clinical data and PCCT data. Finally, ablation studies are designed to verify the validity of this study.
\subsection{Datasets}
\subsubsection{Mayo dataset}{The Mayo dataset \cite{moen2021low} from AAPM Grand Challenge was used to validate the algorithms, which consisted of the NDCT data and the corresponding simulated LDCT data from 10 patients. In this study, in order to obtain data at different doses, a simulated LDCT method similar to the study of \cite{zeng2015simple} was used. It can be described as:
\begin{equation}
			\begin{aligned}
				y_{ld}=\ln\frac{I_0}{\mathrm{Poisson}(I_0\exp(-y_{0}))+\mathrm{Gaussian}(0,\sigma_e^2)}, \label{eq21}
			\end{aligned}
\end{equation}
where fan-beam projections $y_{ld}$ and $y_{0}$ represent low-dose and full-dose CT data. An incident photon count $I_0=1.5\times{10}^5$ and electronic noise variance $\sigma_e^2=10$ were used for simulation. The scanning geometry featured a source-to-rotation center distance of $1361.2$ $mm$ and detector-to-rotation center distance of $615.18$ $mm$, with $720$ views acquired over 360-degree. Images were reconstructed via FBP at $512\times512$ resolution. Datasets were simulated at $50\%$, $25\%$, and $10\%$ dose levels, with $8$ cases used for training and $2$ cases ($L067, L096$) reserved for testing.}
\begin{figure}[!htbp]
		\centerline{\includegraphics[width=\columnwidth]{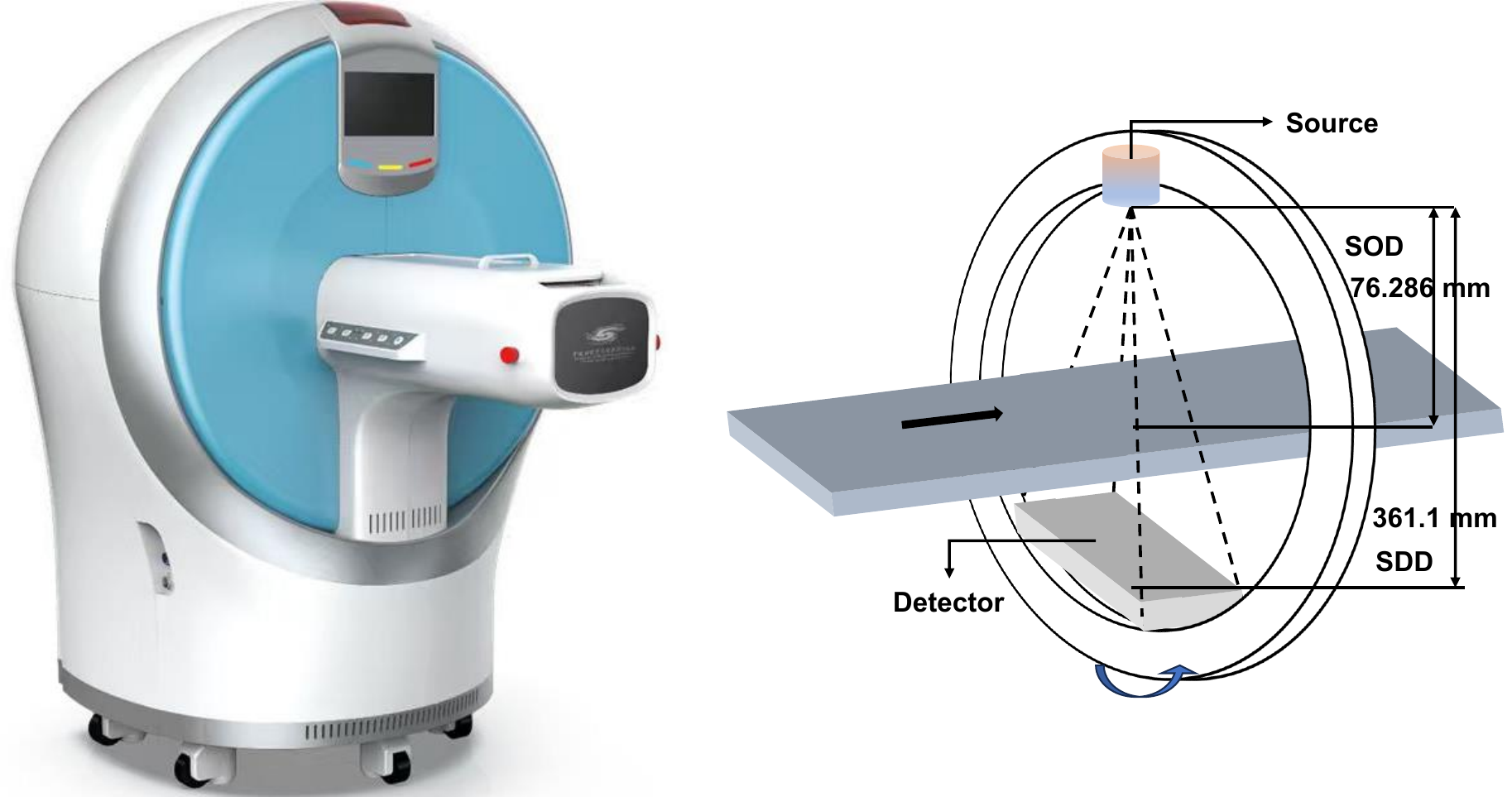}}
		\caption{Spectral CT Imaging System for Small Animals $\mathrm{\mu}$Color SA.}
		\label{fig15}
\end{figure}
\begin{figure*}[!h]
		\centerline{\includegraphics[width=\linewidth]{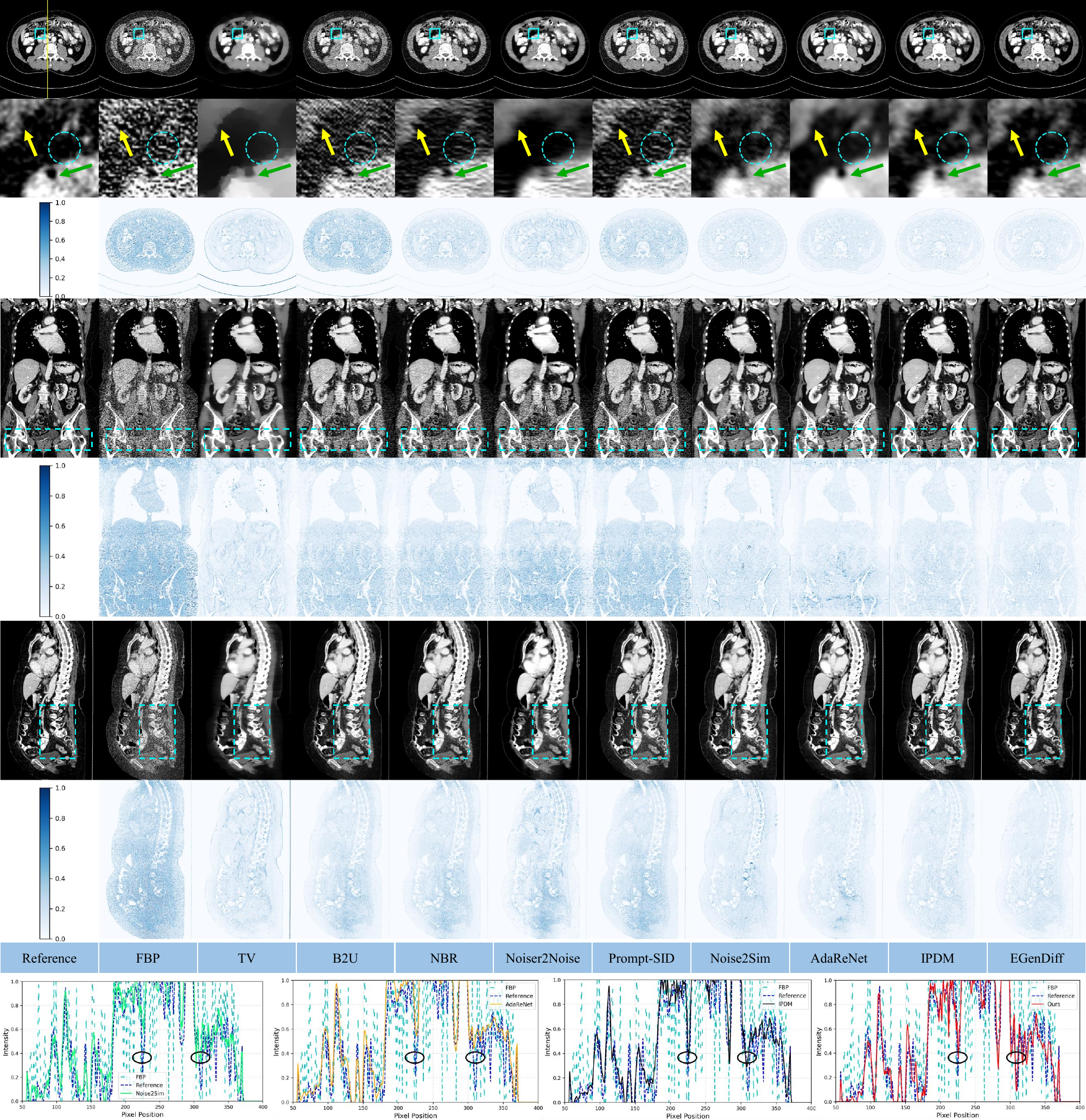}}
		\caption{Qualitative results from the Mayo dataset of $25\%$ dose CT images. A comprehensive qualitative evaluation was conducted on the Mayo $25\%$ dose dataset across transverse, coronal, and sagittal planes. The line charts illustrate the fitting performance of three methods—Noise2Sim, AdaReNet, and IPDM, which produced results comparable to EGenDiff—against reference values for a specific column in the transverse plane. The display window is set to [-100, 200] HU.}
		\label{fig5}
\end{figure*}
\subsubsection{LIDC-IDRI dataset}{To further validate the model, this study utilized the publicly available dataset LIDC-IDRI (Lung Image Database Consortium and Image Database Resource Initiative) \cite{armato2011lung}. It is a publicly available lung nodule dataset, which is mainly used to study early cancer detection in high-risk populations. The dataset contains a total of $1018$ lung CT scan images from $1010$ lung patients. After preprocessing, a simulation approach similar to that of the Mayo dataset was taken to derive $25\%$, $10\%$ dose data. Finally, $45$ patients were selected for training and $5$ patients for testing.}
\subsubsection{Real clinical dataset}{The real clinical data contained a total of $500$ LDCT projection data. These data were acquired using a GE scanner (Discovery CT750 HD) at $120$ $kVp$, $80$ $mAs$ and $1.25$ $mm$ slice thickness. The work utilized only real data for testing, aiming to validate the scalability of generalization under realistic noise and low-energy spectrum conditions. This clinical data comes from publicly available low-dose data from LIDC-IDRI \cite{armato2011lung}.}
\subsubsection{Mouse dual-energy photon counting CT dataset}{Conducted jointly with the Institute of High Energy Physics (IHEP), Chinese Academy of Sciences, this work examined how EGenDiff can be extended to PCCT imaging. As shown in Fig. \ref{fig15}, dual-energy mouse CT data were acquired using the $\mu$Color SA system developed by the IHEP, Chinese Academy of Sciences. Each scan acquired $2000$ projections with a dimension of $2000\times252$, covering energy ranges of $21–30$ $keV$ and $30–70$ $keV$. Scanning parameters included: source-to-object distance $76.286$ $mm$, source-to-detector distance $361.1$ $mm$, and pixel size $0.1$ $mm$. The device uses an advanced photon counting detector with a spatial resolution of up to 15 $\mu m$. It is mainly used for live animal CT imaging in preclinical experimental studies, and can now be applied to whole-body structural energy spectrum imaging of various rodents. PCCT is inherently challenging yet beneficial: It enables precise utilization of both low-energy and high-energy spectral data. Specifically, low-energy scans provide high contrast and improved material differentiation, whereas high-energy data help mitigate beam hardening and metal artifacts. Since optimizing both energy channels is essential for high-accuracy material decomposition, and given that the structural characteristics of low-energy data often resemble the high-frequency noise artifacts found in conventional LDCT, this dataset was selected to validate the extendable generalization capability of EGenDiff. This experimental procedure complied with animal ethics guidelines and was approved by the Ethics Committee of Nanchang University (Approval No.: 20220726008).}
\begin{figure*}[!h]
		\centerline{\includegraphics[width=\linewidth]{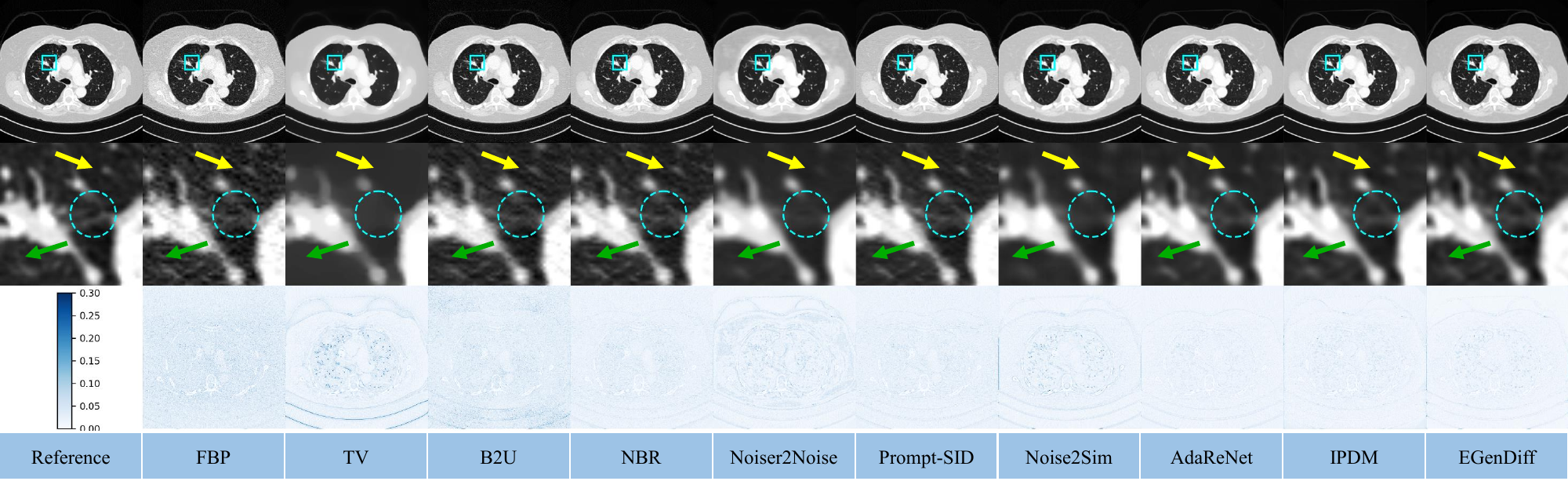}}
		\caption{Qualitative assessment in the lungs against the Mayo $25\%$ dose dataset. The blue region of interest is zoomed in to show the difference, and the green and yellow arrows point to lesions with significant disparity. The display window is [-1350, 200] HU.}
		\label{fig6}
\end{figure*}
\begin{figure*}[!h]
		\centerline{\includegraphics[width=\linewidth]{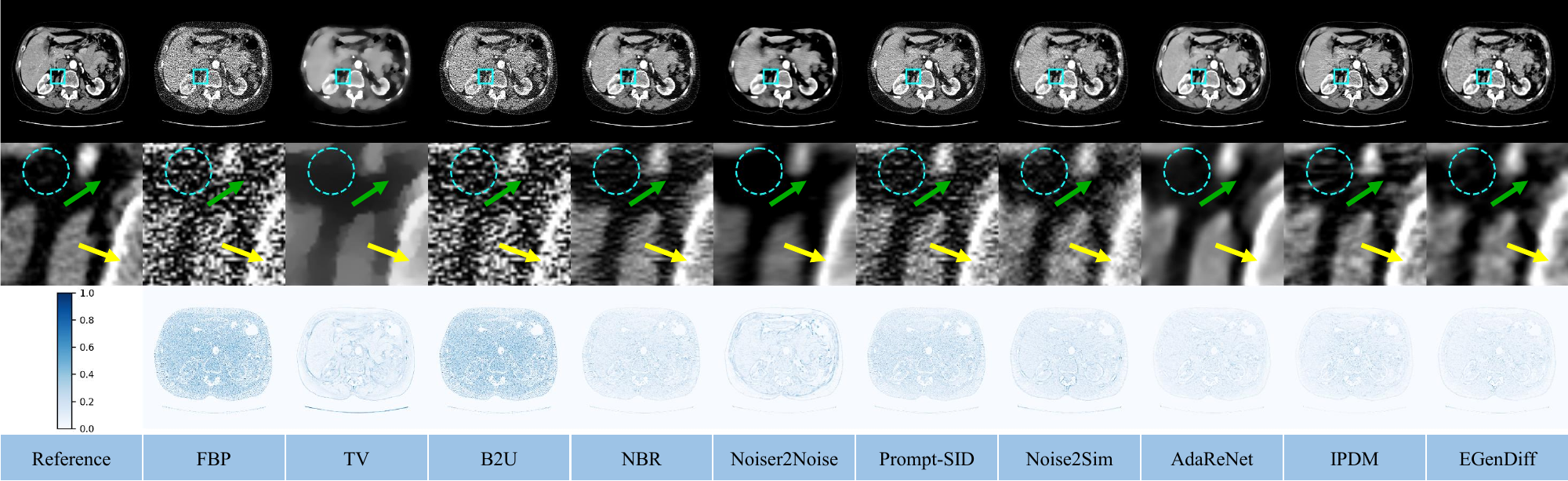}}
		\caption{Qualitative assessment against the LIDC-IDRI $25\%$ dose dataset in the abdomen. The blue region of interest is zoomed in to show the difference, and the green and yellow arrows point to lesions with significant disparity. The display window is [-100, 200].}
		\label{fig7}
\end{figure*}
\subsection{Implementation Details}
The contextual sub-data projection domain model adopts a U-Net architecture similar to \cite{laine2019high}, though with two downsampling stages removed to accommodate projection domain dimensions. Training employs an Adam optimizer with initial learning rate $0.0003$, batch size $8$, and $100$ epochs, with learning rate halved every $20$ epochs. The latent diffusion module utilizes a four-layer autoencoder and a four-layer Transformer-based denoising network. Diffusion training uses AdamW optimizer with initial learning rate $0.0003$, cosine annealing scheduler, $T=1000$ diffusion steps, $T_L=5$ sampling steps, batch size $4$, and $60$ epochs. All experiments were implemented in PyTorch on an NVIDIA RTX 4090D GPU (24GB), requiring approximately $5$ hours for projection domain training and $20$ hours for image domain training. Source code is made available at \href{https://github.com/yqx7150/EGenDiff}{https://github.com/yqx7150/EGenDiff}.
	
In order to evaluate the state-of-the-art of this study, we selected traditional denoising methods and several self-supervised/unsupervised methods for comparison. These include Total Variation (TV) \cite{rudin1992nonlinear}, B2U \cite{wang2022blind2unblind}, Neighbor2Neighbor (NBR) \cite{huang2022neighbor2neighbor}, Noiser2Noise \cite{moran2020noisier2noise}, Prompt-SID \cite{li2025prompt}, Noise2Sim \cite{niu2022noise}, AdaReNet \cite{liu2025rotation}, and IPDM \cite{liao2024domain}.
\subsection{Evaluation Metrics}
To validate the clinical feasibility of the method, three other metrics that are close to the radiologist's judgment of the condition were introduced in this study: feature similarity index (FSIM) \cite{zhang2011fsim}, visual information fidelity (VIF) \cite{sheikh2006image}, and noise quality metric (NQM) \cite{damera2000image}. All metrics were calculated under the CT window [-1024, 3072] HU.
\begin{table*}
	\centering
	\setlength{\extrarowheight}{0pt}
	\addtolength{\extrarowheight}{\aboverulesep}
	\addtolength{\extrarowheight}{\belowrulesep}
	\setlength{\aboverulesep}{0pt}
	\setlength{\belowrulesep}{0pt}
	\caption{Quantitative generalization results (mean±std) on different doses level from the Mayo 2016 dataset and the LIDC-IDRI dataset. The best results are highlighted in bold.}
	\label{table2}
	\resizebox{\linewidth}{!}{%
		\begin{tabular}{ccccccc|cccccc} 
			\toprule
			\multirow{3}{*}{Methods}                       & \multicolumn{3}{c}{Mayo Dataset}                                & \multicolumn{3}{c|}{Mayo Dataset}                               & \multicolumn{3}{c}{LIDC-IDRI Dataset}                            & \multicolumn{3}{c}{LIDC-IDRI Dataset $\to$ Mayo Dataset}         \\
			& \multicolumn{3}{c}{25\% $\to$ 10\%}                             & \multicolumn{3}{c|}{25\% $\to$ 50\%}                            & \multicolumn{3}{c}{25\% $\to$ 10\%}                              & \multicolumn{3}{c}{25\% $\to$ 50\%}                              \\
			& PSNR$\uparrow$      & SSIM(\%) $\uparrow$ & RMSE $\downarrow$   & PSNR $\uparrow$     & SSIM(\%) $\uparrow$ & RMSE $\downarrow$   & PSNR $\uparrow$     & SSIM(\%) $\uparrow$ & RMSE $\downarrow$    & PSNR $\uparrow$     & SSIM(\%) $\uparrow$ & RMSE $\downarrow$    \\ 
			\hline
			\textbf{Baseline/Tradition}                    & \multicolumn{12}{l}{}                                                                                                                                                                                                                                                   \\ 
			\hline
			\rowcolor[rgb]{0.753,0.753,0.753} FBP          & 26.60±4.04          & 49.79±13.28         & 105.38±58.21        & 34.98±2.26          & 81.44±7.24          & 36.84±9.60          & 32.20±4.33          & 70.79±15.97         & 55.93±30.78          & 34.98±2.26          & 81.44±7.24          & 36.84±9.60           \\
			TV                                             & 38.21±2.47          & 94.30±2.49          & 24.74±9.21          & 39.41±1.20          & 85.24±6.20          & 24.83±2.68          & 39.05±1.21          & 80.03±0.34          & 40.03±3.61           & 39.41±1.20          & 85.24±6.20          & 24.83±2.68           \\ 
			\hline
			\textbf{Self-supervised}                       & \multicolumn{12}{l}{}                                                                                                                                                                                                                                                   \\ 
			\hline
			\rowcolor[rgb]{0.753,0.753,0.753} B2U          & 29.94±4.51          & 68.75±12.36         & 73.50±43.38         & 37.36±2.24          & 89.23±5.01          & 28.03±7.55          & 34.56±4.52          & 81.32±11.77         & 43.59±28.67          & 36.93±2.22          & 88.20±5.35          & 29.44±7.84           \\
			Neighbor2Neighbor                              & 34.13±4.67          & 82.13±10.58         & 46.30±30.85         & 41.13±1.67          & 95.59±1.83          & 17.88±3.62          & 38.31±4.28          & 91.06±7.06          & 28.31±20.77          & 40.62±1.75          & 94.95±2.24          & 19.00±4.06           \\
			\rowcolor[rgb]{0.753,0.753,0.753} Noiser2Noise & 32.11±6.52          & 77.39±15.33         & 67.33±59.56         & 39.68±0.84          & 95.78±1.15          & 20.83±2.04          & 37.03±3.80          & 92.04±7.17          & 32.22±23.99          & 36.70±1.13          & 95.46±0.96          & 29.48±3.87           \\
			Prompt-SID                                     & 31.81±4.36          & 73.28±11.89         & 58.94±35.08         & 39.21±1.92          & 92.72±3.34          & 22.44±5.15          & 36.75±4.43          & 86.61±9.45          & 33.83±22.89          & 39.23±1.93          & 92.78±3.35          & 22.41±5.20           \\
			\rowcolor[rgb]{0.753,0.753,0.753} Noise2Sim    & 36.14±4.67          & 90.99±6.54          & 37.92±32.10         & 39.86±1.36          & 96.31±0.70          & 20.58±.56           & 35.73±3.17          & 93.91±3.35          & 35.61±19.40          & 36.24±2.42          & 95.51±1.10          & 32.04±9.05           \\
			AdaReNet                                       & 33.29±9.36          & 86.26±13.66         & 81.95±13.66         & 42.47±1.12          & 97.43±0.56          & 15.17±2.06          & 40.98±3.91          & 96.23±4.53          & \textbf{20.31±16.27} & \textbf{42.13±0.64} & 97.20±0.48          & 16.99±1.21           \\ 
			\hline
			\textbf{Unsupervised}                          & \multicolumn{12}{l}{}                                                                                                                                                                                                                                                   \\ 
			\hline
			\rowcolor[rgb]{0.753,0.753,0.753} IPDM         & 38.02±4.27          & 91.46±9.27          & 29.53±22.78         & 41.93±0.76          & 97.35±0.52          & 16.07±1.43          & 40.59±3.10          & 95.73±3.83          & 21.30±11.05          & 41.57±0.70          & 97.18±0.61          & 16.73±1.39           \\ 
			\hline
			\textbf{\textbf{Self-supervised}}              & \multicolumn{12}{l}{}                                                                                                                                                                                                                                                   \\ 
			\hline
			EGenDiff (\textbf{ours})                       & \textbf{40.82±1.44} & \textbf{96.13±1.11} & \textbf{18.44±3.30} & \textbf{43.11±1.10} & \textbf{97.46±0.53} & \textbf{14.09±1.85} & \textbf{41.14±4.02} & \textbf{96.26±4.01} & 20.95±20.06          & 41.85±0.98          & \textbf{97.40±0.52} & \textbf{16.25±1.89}  \\
			\bottomrule
		\end{tabular}
	}
\end{table*}
\subsection{Experiment Results}
\subsubsection{Evaluation of different datasets}{This study quantitatively validates the model's superiority using $25\%$ dose data from the Mayo and LIDC-IDRI datasets. Comparative experiments against traditional methods, self-supervised approaches, and clean-data-based iterative diffusion methods demonstrate the advancement of our approach from multiple perspectives, as clearly shown in Fig. \ref{fig5} through transverse images, coronal images, sagittal images, zoomed regions, and residual maps. While traditional TV regularization achieves noise removal, it causes over-smoothing and compromises pathological features, with parameter sensitivity across different scanning conditions. Deep learning-based B2U suffers performance degradation due to limited adaptability to structured CT noise. Noiser2Noise introduces excessive randomness by adding artificial noise, leading to texture loss similar to TV methods. Prompt-SID, despite incorporating diffusion models and cue learning, shows limited effectiveness on the Mayo dataset. NBR, lacking our similarity enhancement strategy, produces less distinct lesion visualization in both lung and abdominal scans, as evidenced in Fig. \ref{fig5} and Fig. \ref{fig6}.
\begin{figure}[!h]
			\centerline{\includegraphics[width=\linewidth]{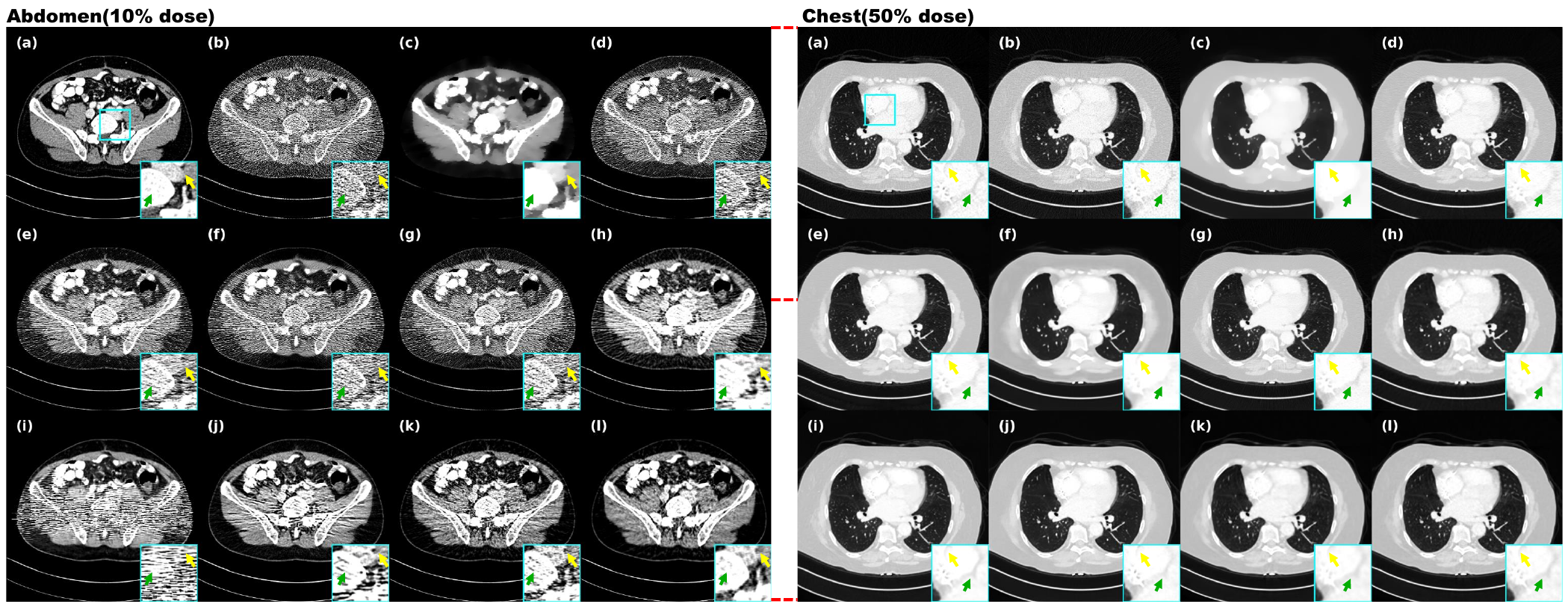}}
			\caption{Display of results on higher and lower dose generalization for (a) Reference (b) FBP (c) TV (d) B2U (e) NBR (f) Noiser2Noise (g) Prompt-SID (h) Noise2Sim (i) AdaReNet (j) IPDM (k) EGenDiff-Higher/Lower and (l) EGenDiff (\textbf{ours}). The display window for abdomen images is [-100, 200] HU and for chest images is [-1350, 200] HU.}
			\label{fig9}
\end{figure}
\begin{figure}[!h]
			\centerline{\includegraphics[width=\linewidth]{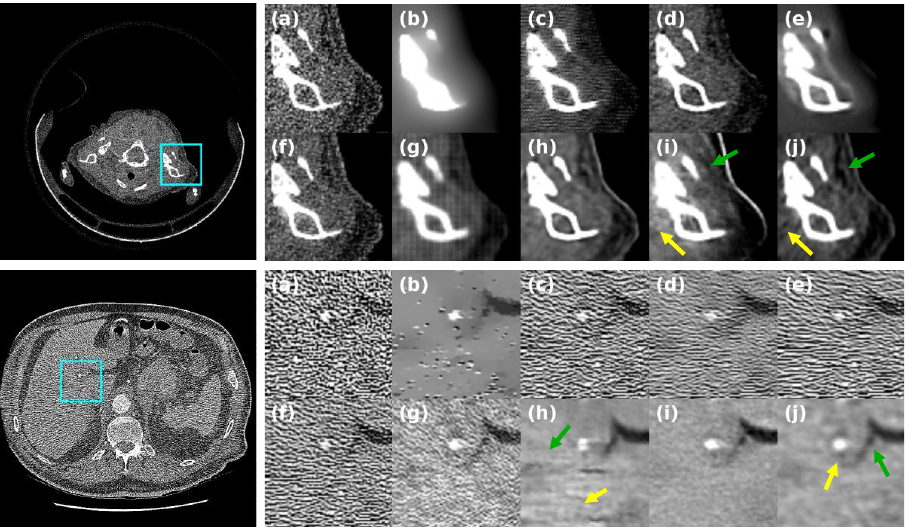}}
			\caption{Generalization results of mouse dual-energy CT and real clinical data. (a) FBP (b) TV (c) B2U (d) NBR (e) Noiser2Noise (f) Prompt-SID (g) Noise2Sim (h) AdaReNet (i) IPDM (j) EGenDiff (\textbf{ours}). The display window is [-100, 200] HU.}
			\label{fig10}
\end{figure}
\begin{figure}[!h]
			\centerline{\includegraphics[width=\linewidth]{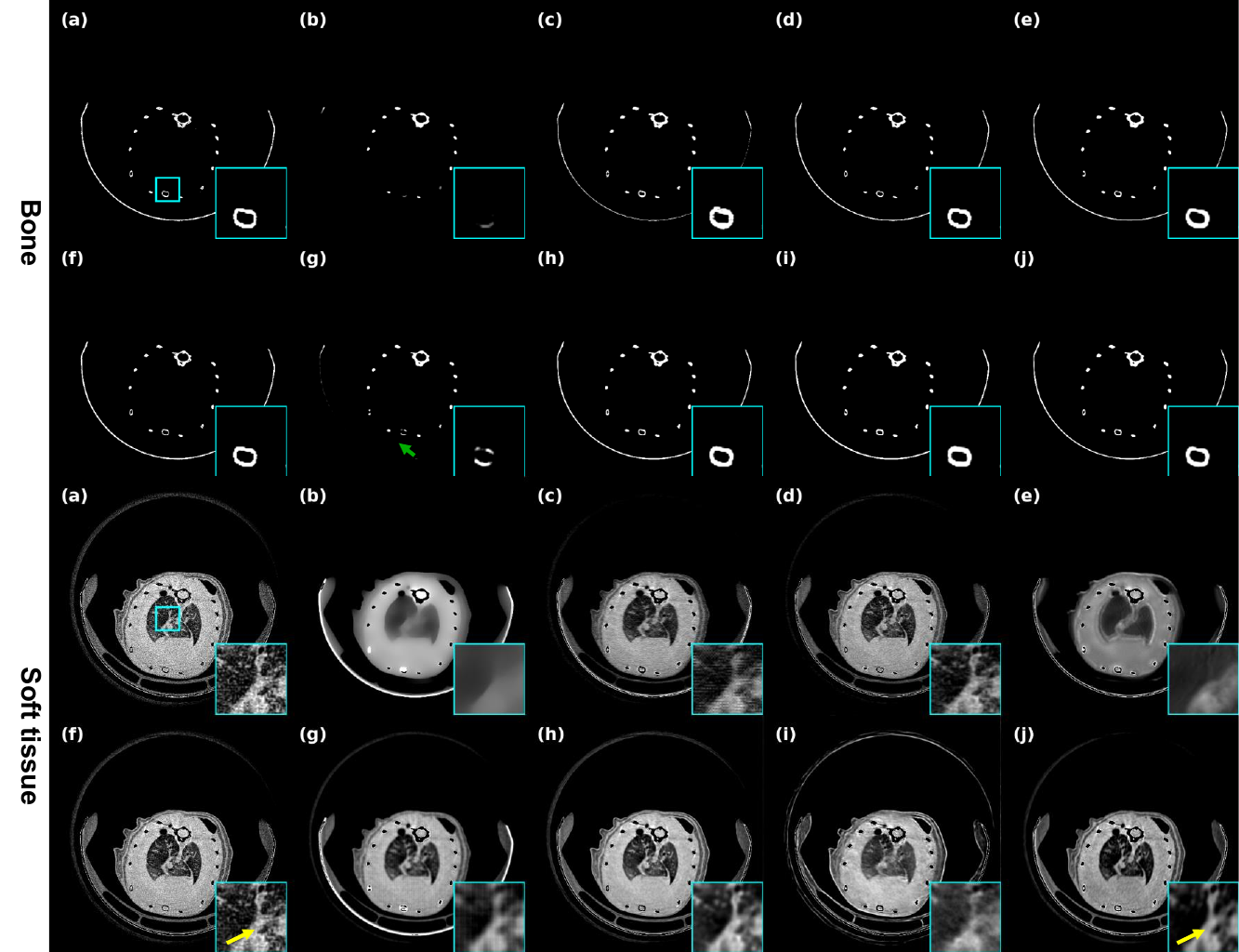}}
			\caption{Results of material decomposition in mouse dual-energy CT data. (a) FBP (b) TV (c) B2U (d) NBR (e) Noiser2Noise (f) Prompt-SID (g) Noise2Sim (h) AdaReNet (i) IPDM (j) EGenDiff (\textbf{ours}). The display window is [100, 400] HU.}
			\label{fig16}
\end{figure}
\begin{table*}[!htbp]
			\centering
			\caption{Ablation studies (mean±std) on the proposed modules and $\lambda$ on 25\% dose test data from the Mayo dataset.}
			\label{table3}
			\resizebox{\linewidth}{!}{%
				\begin{tabular}{ccccc|cccc|cccc} 
					\toprule
					Proj.        & Imag.        & PSNR $\uparrow$              & SSIM(\%) $\uparrow$          & RMSE $\downarrow$            & Dissimilarity        & PSNR~$\uparrow$                                & SSIM(\%)~$\uparrow$                            & RMSE~$\downarrow$                                                                  & $\lambda$                  & PSNR~$\uparrow$              & SSIM(\%)~$\uparrow$          & RMSE~$\downarrow$             \\ 
					\hline
					$\checkmark$ & $\times$     & 41.11±1.07                   & 95.52±1.03                   & 17.73±2.28                   & $\checkmark$         & 41.09±0.87~                                    & 95.46±0.62~                                    & 17.52±1.79~~                                                                       & $\lambda = 0.5$            & 42.21±1.09                   & 96.88±0.66                   & 15.61±2.03                    \\
					&              &                              &                              &                              & \multicolumn{1}{l}{} & \multicolumn{1}{l}{}                           & \multicolumn{1}{l}{}                           & \multicolumn{1}{l|}{}                                                              & $\lambda = 1$              & 42.28±1.24                   & 96.79±0.95                   & 15.54±2.33                    \\
					$\checkmark$ & $\checkmark$ & \textbf{\textbf{42.41±1.21}} & \textbf{\textbf{96.90±2.49}} & \textbf{\textbf{15.30±2.21}} & $\times$             & \textbf{\textbf{\textbf{\textbf{41.11±1.07}}}} & \textbf{\textbf{\textbf{\textbf{95.52±1.03}}}} & \textbf{\textbf{\textbf{\textbf{\textbf{\textbf{\textbf{\textbf{17.73±2.28}}}}}}}} & ~ $\lambda$(\textbf{ours}) & \textbf{\textbf{42.41±1.21}} & \textbf{\textbf{96.90±2.49}} & \textbf{\textbf{15.30±2.21}}  \\
					\bottomrule
				\end{tabular}
			}
\end{table*}
		
As can be seen from Table \ref{table1}, Noise2Sim, AdaReNet, and IPDM all showed good results, but Noise2Sim's method based on batch data similarity requires higher data z-axis orientation, which resulted in a significant drop in metrics across different datasets, and the overly randomized selection of similar data can easily lead to drifting of CT values as shown in Figs. \ref{fig5} and \ref{fig7}. However, our method mitigates the CT value mismatch due to the pixel-level weighted correction that achieves a balance in suppressing noise and preserving details. Notably, although the effect on the LIDC data is lower than that of AdaReNet, it requires independent noisy data with the same information, which cannot be strictly classified as self-supervised, and as shown in Fig. \ref{fig7}, the image produces a smoothing of the details on the muscle tissues and bones, which can interfere with the diagnostic performance of the radiologist to a large extent.
		
Our method best matches the reference in intensity variation, as shown in the line profile in Fig. \ref{fig5}, demonstrating superior reconstruction robustness. Crucially, our approach operates solely on single LDCT projection during both training and inference, offering a truly self-supervised solution suitable for label-scarce clinical environments.}
\subsubsection{Experiment results of extendable generalization}{While self-supervised methods typically suffer from degraded generalization when pre-trained models are applied to different dose levels, this study demonstrates the feasibility and superior generalization capability of our approach across varying doses. As shown in Fig. \ref{fig9} and Table \ref{table2}, our method achieves a breakthrough lead in both quantitative metrics and visual quality for $10\%$ dose abdominal data. Although AdaReNet performs slightly better on lung data, its performance degrades significantly on abdomen data, while our model effectively suppresses scattering artifacts—a capability not possessed by other methods. In addition, the difference between the higher and lower dose strategies exhibited in Fig. \ref{fig9} is a good indication of the correctness of the generalization strategy for scalability. Inspired by NEED, we also generalize the LIDC-IDRI pre-trained model to the Mayo dataset, confirming its reconstruction capability on unseen data.}
\subsubsection{Representation of real clinical data and mouse PCCT data}{To better demonstrate the scalability of generalization of EGenDiff, validation was conducted using both mouse and real clinical data. As demonstrated in Fig. \ref{fig10}, our method outperforms AdaReNet in detail recovery and noise suppression. It effectively removes noise around lesions while preserving diagnostic features—capabilities crucial for clinical practice. IPDM exhibits smoothing and detail loss after dual-domain processing, while Noise2Sim obscures structurally important regions. Though other methods surpass FBP, they show limited adaptability to real noise patterns. On the contrary, EGenDiff removes the noise to a large extent and the pathological features are well reconstructed. Meanwhile, the material decomposition results of mouse spectral CT shown in Fig. \ref{fig16} also indicate that EGenDiff can be extended to different fields and achieve the best results. It will bring a new paradigm to self-supervised denoising methods.}
\begin{figure}[!htbp]
		\centerline{\includegraphics[width=\columnwidth]{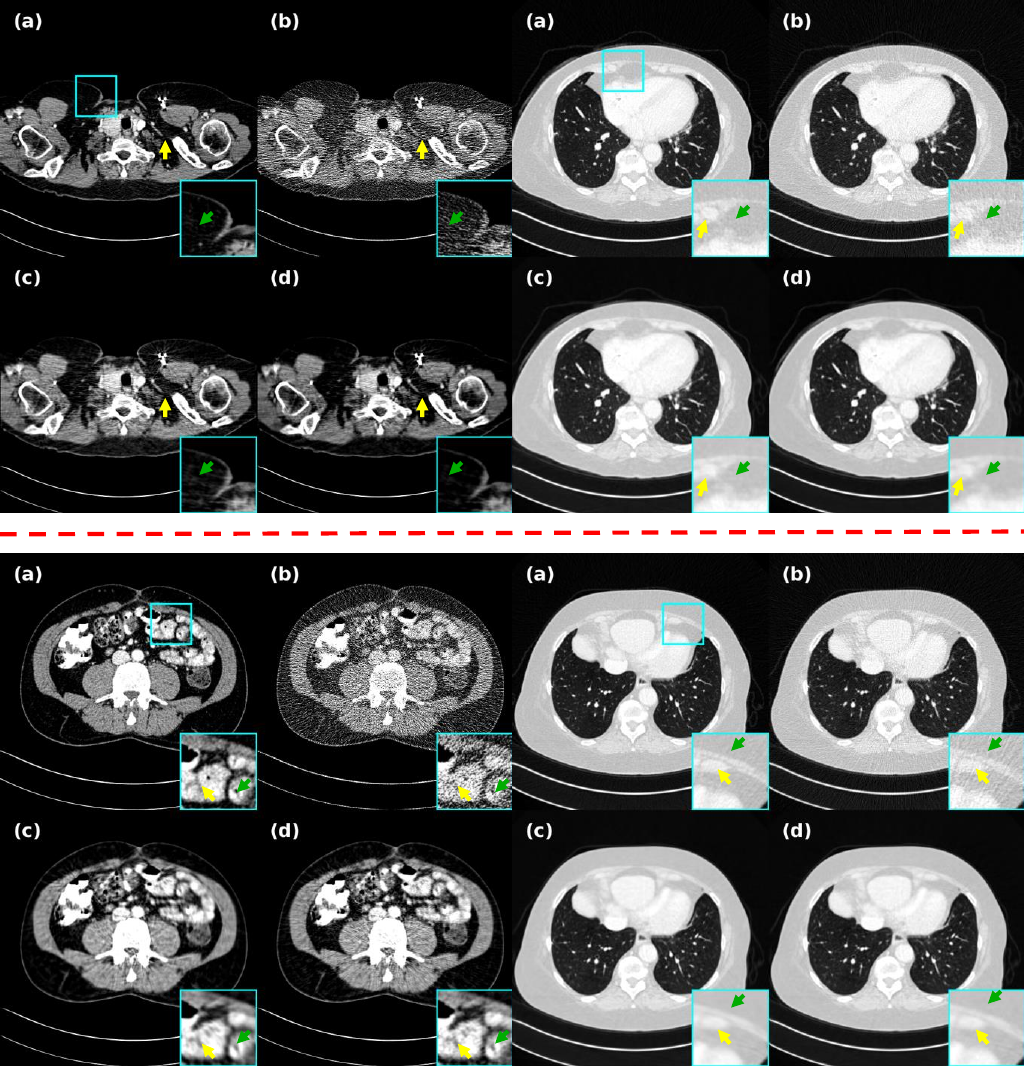}}
		\caption{Ablation studies to verify the effectiveness of two-stage and the importance of similarity enhancement perception. Above: (a) Reference (b) FBP (c) Only projection (d) EGenDiff (\textbf{ours}). Below: (c) No enhance. The display window for abdomen images is [-100, 200] HU and for chest images is [-1350, 200] HU.}
		\label{fig11}
\end{figure}
\subsection{Ablation Study}
\subsubsection{Validity of the dual domain strategy}{This study first presents a holistic analysis of the proposed framework and performs ablation studies on single projection and overall stages. As shown in Fig. \ref{fig11} and Table \ref{table3}, denoising solely in the projection domain reveals inadequate noise suppression, incomplete recovery of muscle tissue details, and suboptimal restoration of skeletal structures. These limitations are effectively addressed through subsequent knowledge distillation in the image domain, demonstrating that the initial reconstruction ${\hat{x}}_0$ and original data $x_{ld}$ serve as complementary guidance, compensating for each other's limitations during the reconstruction process.}
\begin{figure}[!htbp]
		\centerline{\includegraphics[width=\columnwidth]{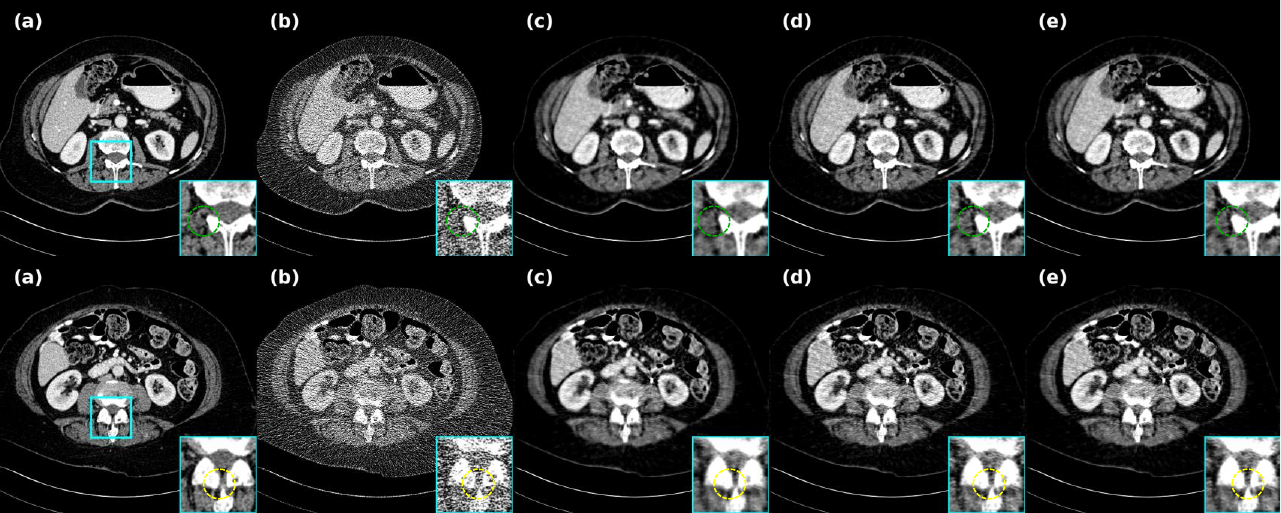}}
		\caption{Ablation studies on the effectiveness of pixel-level weighted fusion. (a) Reference (b) FBP (c) $\lambda = 0.5$ (d) $\lambda = 1$ (e) EGenDiff (\textbf{ours}). The display window for abdomen images is [-100, 200] HU.}
		\label{fig13}
\end{figure}
\subsubsection{Importance of self-enhancing similarity in projection domain}{After analyzing the effectiveness of the dual-domain phase, we focus on the proposed self-enhancing similarity strategy in the projection domain phase. This strategy is proposed to shorten the dissimilarity between contextual sub-data. So it focuses on highlighting the loss weight of similarity and ignoring the excessive part of dissimilarity. As shown in Fig. \ref{fig11} and Table \ref{table3}, implementations without this strategy exhibit feature smoothing and gradient disappearance, impairing the subsequent image-domain refinement stage's ability to discriminate between noise and tissue structures, ultimately leading to suboptimal performance.}
\subsubsection{Pixel-level self-correcting validity and difference from fixed weights}{Subsequently, we carried out the difference between self-correction at the pixel level of the inference process and taking fixed weights. Fig. \ref{fig13} and Table \ref{table3} add fully illustrate the situation, with fixed weights we selected $\lambda$ as $0.5$, and even selected $\lambda$ as $1$, even though this is not in line with the original intent of the dual-stage. As can be seen, our pixel-level fusion can make a balance between ${\hat{x}}_0$ and $x_{ld}$ for each pixel, laying the groundwork for entering into image inference, and showing competitive results in both metrics and reconstruction.}
\section{Discussion}
This study investigates a solution for LDCT reconstruction that relies only on single-dose LDCT projection data, requiring neither paired nor clean data. This is a critical challenge in both deep learning and clinical diagnostics. While our method demonstrates promising results, some reconstructed lung and abdominal images still exhibit incomplete recovery of fine vasculature and subtle blurring in localized pathological regions. We attribute this limitation to substantial corruption of such details by noise, which cannot be fully compensated without clean data-driven distribution modeling. Meanwhile, the value of $N_{x_{ld}}^{25\%}$ varies in different scanning instruments. This paper only studies the feasibility of extendable generalization. Future work will incorporate multi-device physics factors into the reconstruction process, aiming to achieve stronger generalization performance even in the absence of target data.
\section{Conclusion}
In this study, we proposed a novel self-supervised method, EGenDiff, for low-dose CT denoising, and it was also the first self-supervised LDCT denoising method that combines the diffusion model. The main contributions of EGenDiff were 1) proposed a self-enhancing similarity strategy for contextual sub-data, which mitigated image smoothing and provided the initial a priori as a foundation for the follow-up. 2) Combined the initial a priori with latent diffusion depth in the form of knowledge distillation. 3) The pixel-level self-correcting fusion technique in the inference process enabled the fusion of the initial a priori with the balanced LDCT image, and the technique also provided support for generalizing the higher doses, lower doses, doses of unseen data, and PCCT data. 4) EGenDiff enabled highly efficient reconstruction of multiple doses with strong generalization by training only on single-dose data. Both training and testing of EGenDiff only required the LDCT projection domain, making it a truly self-supervised denoising method.
\section*{Appendix}
\subsection{Proof of Eq. \eqref{eq11}}
\label{appendixA}
This appendix analyzes each of the two terms in Eq. \eqref{eq11} separately. First, it analyzes how the key variables $\varepsilon_1$ and $\varepsilon_2$ of self-enhancing similarity are integrated into the training process. Then, it explains their effectiveness in eliminating unreliable regions. Finally, we verify the latter term by using regularization constraints to ensure consistency between input and output.
	
Define $s_1(y_{ld})$, $s_2(y_{ld})$, and $s(y_0)$ as the noisy sub-data, noisy sub-target, and clean sub-data. $\varepsilon_1=\mathbb{E}_{s_2(y_{ld})|s_1(y_0)}(s_2(y_{ld}))-s(y_0)$, $w(\varepsilon_1)=1-\varepsilon_1$. Using the above definition for supervised learning, the objective function is:
\begin{equation}
		\begin{aligned}
			& \mathbb{E}_{y_0,y_{ld}}\parallel w(\varepsilon_1)\odot(f_\theta(s_1(y_{ld}))-s(y_0))\parallel_2^2 \\
			& =\mathbb{E}\parallel w(\varepsilon_1)\odot(f_\theta(s_1(y_{ld}))-s_2(y_{ld})+s_2(y_{ld})-s(y_0))\parallel_2^2, 
		\end{aligned}
		\label{a1}
\end{equation}
under the independent condition that $s\left(y_0\right)$ exists, we expand Eq. \eqref{a1} using the vector norm:
\begin{equation}
		\begin{aligned}
			& \mathbb{E}_{y_0,y_{ld}}\parallel w(\varepsilon_1)\odot (f_\theta(s_1(y_{ld}))-s_2(y_{ld}))\parallel_2^2 \\
			& =\mathbb{E}_{y_{0},y_{ld}}\parallel w(\varepsilon_{1})\odot (f_{\theta}(s_{1}(y_{ld}))-s(y_{0}))\parallel_{2}^{2}+w(\varepsilon_1)^2\sigma_{s_2(y_{ld})}^2 \\
			& -2\varepsilon_{1}\mathbb{E}_{y_{0},y_{ld}}(w(\varepsilon_{1})^2\odot(f_{\theta}(s_{1}(y_{ld}))-s(y_{0}))),
		\end{aligned}
\end{equation}
When $\varepsilon_1\rightarrow0$, the noise pairs exhibit high similarity, effectively approximating the performance of the supervised loss. However, when the similarity regions change significantly, the deviation between Eq. \eqref{eq10} and the supervised error becomes increasingly larger. To alleviate this problem, we introduce $w(\varepsilon_1)$, which eliminates unreliable regions, thus approximating the supervised form not only when $\varepsilon_1\rightarrow0$, but also when $\varepsilon_1\rightarrow1$ (where the context constraints always maintain similarity). $\varepsilon_1$ and $w(\varepsilon_1)$ together act as a smart regulator, thus verifying the first term of Eq. \eqref{eq11}.
	
Defining $\varepsilon_2$ as the output sampling discrepancy. We show that the input difference $\varepsilon_1$ and the output difference jointly constrain denoising consistency; a smaller input difference should correspond to a smaller output difference. The addition of this second term makes training more complete and robust, thus demonstrating the rationale behind the self-enhancing similarity theory.


\begin{thebibliography}{00}
	\bibitem{brenner2007computed}
	D. J. Brenner and E. J. Hall, ``Computed tomography---an increasing source of radiation exposure,'' N. Engl. J. Med., vol. 357, no. 22, pp. 2277--2284, 2007.
	
	\bibitem{smith2009radiation}
	R. Smith-Bindman \emph{et al.}, ``Radiation dose associated with common computed tomography examinations and the associated lifetime attributable risk of cancer,'' Arch. Intern. Med., vol. 169, no. 22, pp. 2078--2086, 2009.
	
	\bibitem{sodickson2009recurrent}
	A. Sodickson \emph{et al.}, ``Recurrent CT, cumulative radiation exposure, and associated radiation-induced cancer risks from CT of adults,'' Radiology, vol. 251, no. 1, pp. 175--184, 2009.
	
	\bibitem{balda2012ray}
	M. Balda, J. Hornegger, and B. Heismann, ``Ray contribution masks for structure adaptive sinogram filtering,'' IEEE Trans. Med. Imaging, vol. 31, no. 6, pp. 1228--1239, 2012.
	
	\bibitem{sidky2008image}
	E. Y. Sidky and X. Pan, ``Image reconstruction in circular cone-beam computed tomography by constrained, total-variation minimization,'' Phys. Med. Biol., vol. 53, no. 17, p. 4777, 2008.
	
	\bibitem{chen2014artifact}
	Y. Chen \emph{et al.}, ``Artifact suppressed dictionary learning for low-dose CT image processing,'' IEEE Trans. Med. Imaging, vol. 33, no. 12, pp. 2271--2292, 2014.
	
	\bibitem{manduca2009projection}
	A. Manduca \emph{et al.}, ``Projection space denoising with bilateral filtering and CT noise modeling for dose reduction in CT,'' Med. Phys., vol. 36, no. 11, pp. 4911--4919, 2009.
	
	\bibitem{wang2006penalized}
	J. Wang \emph{et al.}, ``Penalized weighted least-squares approach to sinogram noise reduction and image reconstruction for low-dose x-ray computed tomography,'' IEEE Trans. Med. Imaging, vol. 25, no. 10, pp. 1272--1283, 2006.
	
	\bibitem{dabov2007image}
	K. Dabov \emph{et al.}, ``Image denoising by sparse 3-D transform-domain collaborative filtering,'' IEEE Trans. Image Process., vol. 16, no. 8, pp. 2080--2095, 2007.
	
	\bibitem{willemink2018photon}
	M. J. Willemink \emph{et al.}, ``Photon-counting CT: technical principles and clinical prospects,'' Radiology, vol. 289, no. 2, pp. 293--312, 2018.
	
	\bibitem{shan2019competitive}
	H. Shan \emph{et al.}, ``Competitive performance of a modularized deep neural network compared to commercial algorithms for low-dose CT image reconstruction,'' Nat. Mach. Intell., vol. 1, no. 6, pp. 269--276, 2019.
	
	\bibitem{wang2020deep}
	G. Wang, J. C. Ye, and B. De Man, ``Deep learning for tomographic image reconstruction,'' Nat. Mach. Intell., vol. 2, no. 12, pp. 737--748, 2020.
	
	\bibitem{shen2022mlf}
	J. Shen \emph{et al.}, ``Mlf-iosc: multi-level fusion network with independent operation search cell for low-dose CT denoising,'' IEEE Trans. Med. Imaging, vol. 42, no. 4, pp. 1145--1158, 2022.
	
	\bibitem{chen2017low}
	H. Chen \emph{et al.}, ``Low-dose CT with a residual encoder-decoder convolutional neural network,'' IEEE Trans. Med. Imaging, vol. 36, no. 12, pp. 2524--2535, 2017.
	
	\bibitem{li2025ddoct}
	L. Li \emph{et al.}, ``DDoCT: Morphology preserved dual-domain joint optimization for fast sparse-view low-dose CT imaging,'' Med. Image Anal., vol. 101, p. 103420, 2025.
	
	\bibitem{lehtinen2018noise2noise}
	J. Lehtinen \emph{et al.}, ``Noise2Noise: Learning image restoration without clean data,'' arXiv preprint arXiv:1803.04189, 2018.
	
	\bibitem{niu2022noise}
	C. Niu \emph{et al.}, ``Noise suppression with similarity-based self-supervised deep learning,'' IEEE Trans. Med. Imaging, vol. 42, no. 6, pp. 1590--1602, 2022.
	
	\bibitem{garber2024image}
	T. Garber and T. Tirer, ``Image restoration by denoising diffusion models with iteratively preconditioned guidance,'' In Proc. IEEE/CVF Conf. Comput. Vis. Pattern Recognit. Workshops, pp. 25245--25254, 2024.
	
	\bibitem{ho2020denoising}
	J. Ho, A. J. Jain, and P. Abbeel, ``Denoising diffusion probabilistic models,'' Adv. Neural Inf. Process. Syst., vol. 33, pp. 6840--6851, 2020.
	
	\bibitem{dong2025}
	S. Dong \emph{et al.}, ``A flow-based truncated denoising diffusion model for super-resolution magnetic resonance spectroscopic imaging,'' Med. Image Anal., vol. 99, p. 103358, 2025.
	
	\bibitem{song2020score}
	Y. Song \emph{et al.}, ``Score-based generative modeling through stochastic differential equations,'' arXiv preprint arXiv:2011.13456, 2020.
	
	\bibitem{hoogeboom2023simple}
	E. Hoogeboom, J. Heek, and T. Salimans, ``simple diffusion: End-to-end diffusion for high resolution images,'' In Int. Conf. Mach. Learn., pp. 13213--13232, 2023.
	
	\bibitem{aali2024ambient}
	A. Aali \emph{et al.}, ``Ambient diffusion posterior sampling: Solving inverse problems with diffusion models trained on corrupted data,'' arXiv preprint arXiv:2403.08728, 2024.
	
	\bibitem{zheng2023dpm}
	K. Zheng \emph{et al.}, ``Dpm-solver-v3: Improved diffusion ode solver with empirical model statistics,'' Adv. Neural Inf. Process. Syst., vol. 36, pp. 55502--55542, 2023.
	
	\bibitem{liao2024domain}
	F. Liao \emph{et al.}, ``Domain progressive low-dose CT Imaging using Iterative Partial Diffusion Model,'' IEEE Trans. Med. Imaging, 2024.
	
	\bibitem{nam2024contrastive}
	H. Nam \emph{et al.}, ``Contrastive denoising score for text-guided latent diffusion image editing,'' In Proc. IEEE/CVF Conf. Comput. Vis. Pattern Recognit. Workshops, pp. 9192--9201, 2024.
	
	\bibitem{hu2023self}
	V. T. Hu \emph{et al.}, ``Self-guided diffusion models,'' In Proc. IEEE/CVF Conf. Comput. Vis. Pattern Recognit. Workshops, pp. 18413--18422, 2023.
	
	\bibitem{batson2019noise2self}
	J. Batson and L. Royer, ``Noise2self: Blind denoising by self-supervision,'' In Int. Conf. Mach. Learn., pp. 524--533, 2019.
	
	\bibitem{moran2020noisier2noise}
	N. Moran \emph{et al.}, ``Noisier2noise: Learning to denoise from unpaired noisy data,'' In Proc. IEEE/CVF Conf. Comput. Vis. Pattern Recognit. Workshops, pp. 12064--12072, 2020.
	
	\bibitem{millard2023theoretical}
	C. Millard \emph{et al.}, ``A theoretical framework for self-supervised MR image reconstruction using sub-sampling via variable density Noisier2Noise,'' IEEE Trans. Comput. Imaging, vol. 9, pp. 707--720, 2023.
	
	\bibitem{yaman2020selfsupervised}
	B. Yaman \emph{et al.}, ``Self-supervised learning of physics-guided reconstruction neural networks without fully sampled reference data,'' Magn. Reson. Med., vol. 84, no. 6, pp. 3172--3191, 2020.
	
	\bibitem{yaman2021zeroshot}
	B. Yaman \emph{et al.}, ``Zero-shot self-supervised learning for MRI reconstruction,'' arXiv preprint arXiv:2102.07737, 2021.
	
	\bibitem{huang2022neighbor2neighbor}
	T. Huang \emph{et al.}, ``Neighbor2neighbor: A self-supervised framework for deep image denoising,'' IEEE Trans. Image Process., vol. 31, pp. 4023--4038, 2022.
	
	\bibitem{wang2024svb}
	Y. Wang \emph{et al.}, ``SVB: Self-Supervised Real CT Denoising via Similarity-Based Visual Blind-Spot Scheme,'' IEEE Trans. Instrum. Meas., 2024.
	
	\bibitem{wu2023unsharp}
	Q. Wu \emph{et al.}, ``Unsharp structure guided filtering for self-supervised low-dose CT imaging,'' IEEE Trans. Med. Imaging, vol. 42, no. 11, pp. 3283--3294, 2023.
	
	\bibitem{unal2024proj2proj}
	M. O. Unal, M. Ertas, and I. Yildirim, ``Proj2Proj: self-supervised low-dose CT reconstruction,'' PeerJ Comput. Sci., vol. 10, p. e1849, 2024.
	
	\bibitem{mansour2023zero-shot}
	Y. Mansour and R. Heckel, ``Zero-shot noise2noise: Efficient image denoising without any data,'' In Proc. IEEE/CVF Conf. Comput. Vis. Pattern Recognit. Workshops, pp. 14018--14027, 2023.
	
	\bibitem{zhao2023sample2sample}
	Y. X. Zhao \emph{et al.}, ``Sample2Sample: An improved self-supervised denoising framework for random noise suppression in distributed acoustic sensing vertical seismic profile data,'' Geophys. J. Int., vol. 232, no. 3, pp. 1515--1532, 2023.
	
	\bibitem{liu2022similarity}
	N. Liu \emph{et al.}, ``Similarity-informed self-learning and its application on seismic image denoising,'' IEEE Trans. Geosci. Remote Sens., vol. 60, pp. 1--13, 2022.
	
	\bibitem{shi2024zero-ig}
	Y. Shi \emph{et al.}, ``ZERO-IG: Zero-shot illumination-guided joint denoising and adaptive enhancement for low-light images,'' In Proc. IEEE/CVF Conf. Comput. Vis. Pattern Recognit. Workshops, pp. 3015--3024, 2024.
	
	\bibitem{aali2025robust}
	A. Aali \emph{et al.}, ``Robust multi-coil MRI reconstruction via self-supervised denoising,'' \emph{Magn. Reson. Med.}, 2025.
	
	\bibitem{kim2021noise2score}
	K. Kim and J. C. Ye, ``Noise2score: tweedie’s approach to self-supervised image denoising without clean images,'' Adv. Neural Inf. Process. Syst., vol. 34, pp. 864--874, 2021.
	
	\bibitem{wu2024mu}
	W. Wu \emph{et al.}, ``Multi-Level Noise Sampling From Single Image for Low-Dose Tomography Reconstruction,'' IEEE J. Biomed. Health Inf., 2024.
	
	\bibitem{gao2023corediff}
	Q. Gao \emph{et al.}, ``Corediff: Contextual error-modulated generalized diffusion model for low-dose CT denoising and generalization,'' IEEE Trans. Med. Imaging, vol. 43, no. 2, pp. 745--759, 2023.
	
	\bibitem{gao2025noise}
	Q. Gao \emph{et al.}, ``Noise-Inspired Diffusion Nodel for Generalizable low-dose CT Reconstruction,'' arXiv preprint arXiv:2506.22012, 2025.
	
	\bibitem{yang2025lfdt}
	B. Yang \emph{et al.}, ``LFDT-Fusion: A latent feature-guided diffusion Transformer model for general image fusion,'' Inf. Fusion, vol. 113, p. 102639, 2025.
	
	\bibitem{li2025prompt}
	H. Li \emph{et al.}, ``Prompt-SID: Learning Structural Representation Prompt via Latent Diffusion for Single Image Denoising,'' In Proc. AAAI Conf. Artif. Intell., vol. 39, no. 5, pp. 4734--4742, 2025.
	
	\bibitem{zhussip2019extending}
	M. Zhussip, S. Soltanayev, and S. Y. Chun, ``Extending stein's unbiased risk estimator to train deep denoisers with correlated pairs of noisy images,'' Adv. Neural Inf. Process. Syst., vol. 32, 2019.
	
	\bibitem{moen2021low}
	T. R. Moen \emph{et al.}, ``Low-dose CT image and projection dataset,'' Med. Phys., vol. 48, no. 2, pp. 902--911, 2021.
	
	\bibitem{zeng2015simple}
	D. Zeng \emph{et al.}, ``A simple low-dose x-ray CT simulation from high-dose scan,'' IEEE Trans. Nucl. Sci., vol. 62, no. 5, pp. 2226--2233, 2025.
	
	\bibitem{armato2011lung}
	S. G. Armato III \emph{et al.}, ``The lung image database consortium (LIDC) and image database resource initiative (IDRI): a completed reference database of lung nodules on CT scans,'' Med. Phys., vol. 38, no. 2, pp. 915--931, 2011.
	
	\bibitem{laine2019high}
	S. Laine \emph{et al.}, ``High-quality self-supervised deep image denoising,'' Adv. Neural Inf. Process. Syst., vol. 32, 2019.
	
	\bibitem{rudin1992nonlinear}
	L. I. Rudin, S. Osher, and E. Fatemi, ``Nonlinear total variation based noise removal algorithms,'' Physica D, vol. 60, no. 1-4, pp. 259--228, 1992.
	
	\bibitem{wang2022blind2unblind}
	Z. Wang \emph{et al.}, ``Blind2unblind: Self-supervised image denoising with visible blind spots,'' In Proc. IEEE/CVF Conf. Comput. Vis. Pattern Recognit. Workshops, pp. 2027--2036, 2022.
	
	\bibitem{liu2025rotation}
	H. Liu \emph{et al.}, ``Rotation-Equivariant Self-Supervised Method in Image Denoising,'' In Proc. IEEE/CVF Conf. Comput. Vis. Pattern Recognit. Workshops, pp. 12720--12730, 2025.
	
	\bibitem{wang2004image}
	Z. Wang, A. C. Bovik, H. R. Sheikh, and E. P. Simoncelli, ``Image quality assessment: from error visibility to structural similarity,'' IEEE Trans. Image Process., vol. 13, no. 4, pp. 600--612, 2004.
	
	\bibitem{zhang2011fsim}
	L. Zhang \emph{et al.}, ``FSIM: A feature similarity index for image quality assessment,'' IEEE Trans. Image Process., vol. 20, no. 8, pp. 2378--2386, 2011.
	
	\bibitem{sheikh2006image}
	H. R. Sheikh and A. C. Bovik, ``Image information and visual quality,'' IEEE Trans. Image Process., vol. 15, no. 2, pp. 430--444, 2006.
	
	\bibitem{damera2000image}
	N. Damera-Venkata \emph{et al.}, ``Image quality assessment based on a degradation model,'' IEEE Trans. Image Process., vol. 9, no. 4, pp. 636--650, 2000.
\end{thebibliography}
\end{document}